%% file: main.tex
\documentclass[journal,comsoc]{IEEEtran}
\usepackage{graphicx}
\usepackage{comment}
\usepackage{amsmath,amssymb} 
\usepackage{color}
\usepackage[normalem]{ulem}

\usepackage{multirow}
\usepackage{booktabs}
\usepackage{url}
\usepackage{footnote}
\usepackage{subcaption}

\usepackage{algorithm}
\usepackage{algorithmic}
\usepackage[T1]{fontenc}

\usepackage[capitalise]{cleveref}
\usepackage{soul}
\ifCLASSINFOpdf
\else
\fi
\usepackage{amsmath}
\usepackage[cmintegrals]{newtxmath}
\begin{document}
%
\title{ResKD: Residual-Guided Knowledge Distillation}
%
%
%

\author{Xuewei~Li$^{*}$, 
        Songyuan~Li$^{*}$, 
        Bourahla~Omar,
        Fei Wu,
        and~Xi~Li
\IEEEcompsocitemizethanks{\IEEEcompsocthanksitem X.~Li, S.~Li, B. Omar, F.~Wu and X.~Li are with  College of Computer Science and Technology, Zhejiang University, Hangzhou 310027, China. \protect
E-mail: \{3150104097,~leizungjyun,~bourahla,~xilizju\}@zju.edu.cn. \protect
}
\thanks{ The first two authors (Xuewei Li and Songyuan Li) contribute equally.}

\thanks{(Correspongding author: Xi Li.)}}

\markboth{IEEE  TRANSACTIONS ON IMAGE PROCESSING}%
{Shell \MakeLowercase{\textit{et al.}}:ResKD: Residual-Guided Knowledge Distillation}
%



\maketitle


\input{sections/abstract.tex}


%
\IEEEpeerreviewmaketitle

\input{sections/intro.tex}
\input{sections/related.tex}

\input{sections/method.tex}

\input{sections/experiment.tex}

\input{sections/analysis.tex}
\input{sections/conclusion.tex}

\bibliographystyle{IEEEtran}
\bibliography{IEEEabrv,bibli}

\begin{IEEEbiography}[{\includegraphics[width=1in,height=1.25in,clip,keepaspectratio]{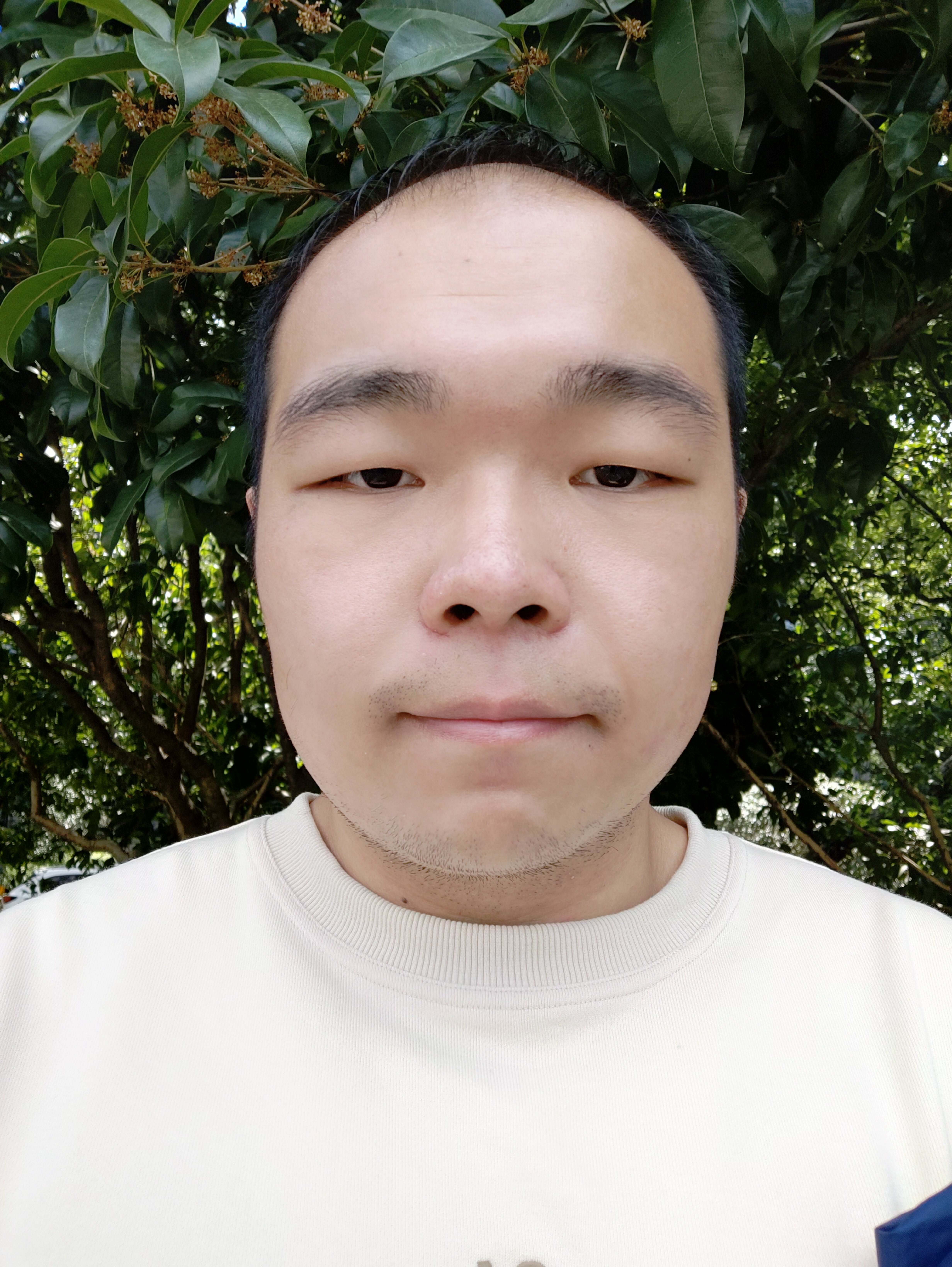}}]{Xuewei Li}
  received his bachelor of engineering in 2019 from Zhejiang University, China. He is currently a Ph.D. candidate at Zhejiang University. His current research interests include knowledge distillation, image classification and visual relationship detection.
\end{IEEEbiography}

\begin{IEEEbiography}[{\includegraphics[width=1in,height=1.25in,clip,keepaspectratio]{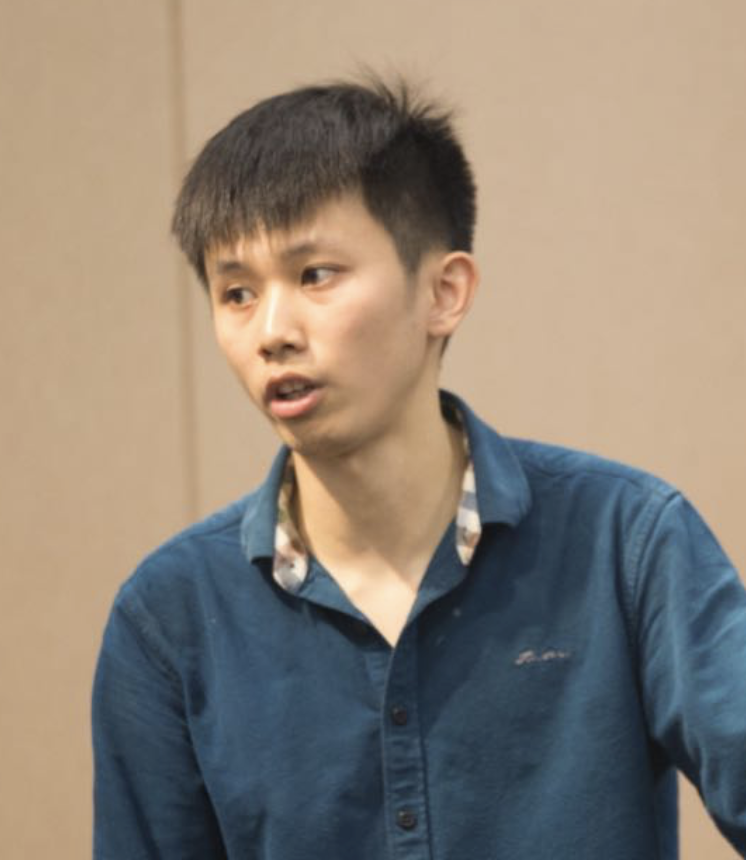}}]{Songyuan Li}
  received his master's degree in 2017 from Zhejiang University, China, where he worked on problems in computer architecture and operating systems. He is currently a Ph.D. candidate at Zhejiang University. His current research interests include knowledge distillation, semantic segmentation and dynamic routing.
\end{IEEEbiography}

\begin{IEEEbiography}[{\includegraphics[width=1in,height=1.25in,clip,keepaspectratio]{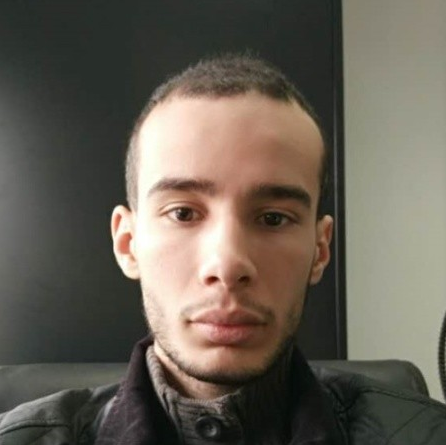}}]{Bourahla Omar}
  received the Master's degree from the University of Science and Technology Houari Boumediene, Algiers, Algeria, in 2015. From 2015 until currenlty (2021), he is pursuing his PhD in Zhejiang University, China. His research interests include machine learning, visual object tracking and rain removal.
\end{IEEEbiography}
\begin{IEEEbiography}[{\includegraphics[width=1in,height=1.25in,clip,keepaspectratio]{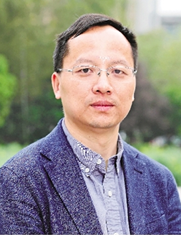}}]{Fei Wu}
  received the B.S. degree from Lanzhou
  University, Lanzhou, Gansu, China, the M.S. degree
  from Macao University, Taipa, Macau, and the Ph.D.
  degree from Zhejiang University, Hangzhou, China.
  He is currently a Full Professor with the College
  of Computer Science and Technology, Zhejiang
  University. He was a Visiting Scholar with Prof.
  B. Yu’s Group, University of California, Berkeley,
  from 2009 to 2010. His current research interests
  include multimedia retrieval, sparse representation,
  and machine learning.
  \end{IEEEbiography}
\begin{IEEEbiography}[{\includegraphics[width=1in,height=1.25in,clip,keepaspectratio]{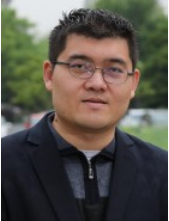}}]{Xi $\textup{Li}^{\dagger}$}
  received the Ph.D. degree from the National Laboratory of Pattern Recognition, Chinese Academy of Sciences, Beijing, China, in 2009. From 2009 to 2010, he was a Post-Doctoral Researcher with CNRS Telecom ParisTech, France. He was a Senior Researcher with the University of Adelaide, Australia. He is currently a Full Professor with Zhejiang University, China. His research interests include visual tracking, compact learning, motion analysis, face recognition, data mining, and image retrieval.
\end{IEEEbiography}

%









\end{document}

%% file: sections/abstract.tex
\begin{abstract}

Knowledge distillation, aimed at transferring the knowledge from a heavy teacher network  to a lightweight student network, has emerged as a promising technique for compressing neural networks. 
However, due to the capacity gap between the heavy teacher and the lightweight student, there still exists a significant performance gap between them. 
In this paper, we see knowledge distillation in a fresh light, using the knowledge gap, or the \emph{residual}, between a teacher and a student as guidance to train a much more lightweight student, called a res-student.  
We combine the student and the res-student into a new student, where the res-student rectifies the errors of the former student. Such a residual-guided process can be repeated until the user strikes the balance between accuracy and cost.
At inference time, we propose a sample-adaptive strategy to decide which res-students are not necessary for each sample, which can save computational cost. Experimental results show that we achieve competitive performance with 18.04$\%$, 23.14$\%$, 53.59$\%$, and 56.86$\%$ of the teachers' computational costs on the CIFAR-10, CIFAR-100, Tiny-ImageNet, and ImageNet datasets.
Finally, we do thorough theoretical and empirical analysis for our method.


\end{abstract}

%% file: sections/intro.tex
\section{Introduction}

\IEEEPARstart{A}{s} deep learning goes deeper, state-of-the-art neural networks~\cite{he2016deep, huang2017densely, tan2019efficientnet, touvron2020fixing} have obtained better and better performance, and yet demand more and more computational resources. 
While models with large capacity can achieve high accuracy, they are impractical for resource-limited devices such as embedded systems. To this end, researchers have studied cost-effective networks~\cite{howard2017mobilenets, sandler2018mobilenetv2, zhang2018shufflenet, ma2018shufflenet} and efficient training strategies~\cite{zhang2018deep, nowak2018deep, crowley2018moonshine, wang2018progressive, liu2019knowledge, gu2020search, chen2019data, dabouei2020supermix, xu2020knowledge}.
Knowledge distillation (KD)~\cite{hinton2015distilling} has emerged as a compression technique where an analogy of the teacher-student relationship is drawn to explain the idea that the knowledge of a powerful yet heavy teacher network can be distilled into a lightweight student network by adding a loss term that encourages the student to mimic the teacher.  

Due to the capacity gap between a heavy teacher and a lightweight student, there is still a significant performance gap between them. 
Existing KD methods have made efforts to modify the loss term to improve the student's performance~\cite{zhang2018deep, peng2019correlation, xie2019self, cho2019efficacy, gou2020knowledge}. 
 However, the discrepancy between a teacher and a student can be considered as knowledge, and it remains relatively unexplored in  knowledge distillation.

In this paper, we see knowledge distillation in a fresh light, using the knowledge gap between a teacher and a student as guidance.
First, we train a student network $S_0$ from a teacher $T$ as usual. 
Then, we train a much more lightweight network, named a res-student, to learn the knowledge gap, or the \emph{residual}, between the teacher $T$ and the student $S_0$. 
The combination of the student $S_0$ and the res-student $R_1$ becomes a new student $S_1$, where $R_1$ corrects the errors of $S_0$. Similarly, an even more lightweight res-student $R_2$ can be used to learn the knowledge gap between $T$ and $S_1$ to build a new student $S_2$.
Such a residual-guided process can be repeated until a final student $S_n$ is obtained.
We call our framework residual-guided knowledge distillation (ResKD).
The idea is akin to approximating functions by a polynomial. In a series expansion of a function, a higher-order polynomial would be a better approximation but would demand more computations.
Similarly, a higher-order ResKD-style student would be closer to the teacher but would be more expensive. Users can control the total capacity of the final student network $S_n$ by setting hyper-parameters for termination.







In addition, the knowledge gap mentioned above is different from sample to sample. For instance, given an $S_2$ student, we observed that, for some images, $S_1$ (i.e., $S_0 + R_1$) or even $S_0$ itself has highly confident scores, while $R_2$ or $R_1+ R_2$ has little contributions. It is unnecessary to use all the res-students for each sample at inference time. Thus, we introduce a sample-adaptive strategy for the inference phase. For each sample, if the confidence of $S_i$ is high enough, we truncate the unnecessary res-students to save computational cost.

We do experiments on several standard benchmarks. 
The experimental results show that we achieve competitive performance with \emph{18.04$\%$, 23.14$\%$, 53.59$\%$, and 56.86$\%$} of the teacher's FLOPs for CIFAR-10, CIFAR-100, Tiny-ImageNet, and ImageNet respectively. 
Also, we apply our ResKD framework to different KD methods and show that the framework is generic to knowledge distillation methods.
Finally, we analyze the effectiveness of this idea and use informativeness~\cite{zhang2019adasample} to visualize the bridging gap process.

Our contributions in this paper are summarized as follows:

\begin{itemize}
\setlength{\itemsep}{0pt}
\setlength{\parsep}{0pt}
\setlength{\parskip}{0pt}
\item We design a residual-guided learning method that uses a series of res-students to  bridge the gap between a student network and a teacher network.
\item We introduce a sample-adaptive strategy at inference time to make our framework adaptive to different samples to save the cost of additional res-student networks. 
\item We evaluate our method on different datasets and perform detailed experiments that showcase the importance of each part of the framework. 
\end{itemize}

%% file: sections/related.tex
\section{Related Work}


\subsection{Knowledge Distillation} 
We categorize knowledge distillation methods in terms of the number of stages.
Traditionally, knowledge distillation is a two-stage method, in which a teacher network  is trained first, and then a student network is trained under the guidance of the teacher network. Bucil\u a et al.~\cite{bucilua2006model} pioneered the idea of transferring the knowledge from a cumbersome model to a small model. 
Hinton et al.~\cite{hinton2015distilling} popularized this idea by the concept of knowledge distillation (KD), in which a student neural network is trained with the benefit of the soft targets provided by teacher networks. 
Compared to traditional one-hot labels, the output from a teacher network contains more information about the fine-grained distribution of data, which helps the student achieve better performance. 
Recently, many works have focused on improving the information propagation way or putting strictness to the distillation process via optimization~\cite{romero2014fitnets, han2019neural, zagoruyko2016paying, peng2019correlation, tung2019similarity, yim2017gift, park2019relational, yuan2020revisiting, ahn2019variational, huang2017like, passalis2018learning, kim2018paraphrasing, heo2019knowledge, tian2019contrastive} to teach the student better. 
For example, Peng et al.~\cite{peng2019correlation} proposed that a student network should not only focus on mimicking from a teacher at an instance level, but also imitating the embedding space of a teacher so that the student can possess intra-class compactness and inter-class separability. 
In addition, the effect of different teachers is also researched~\cite{ba2014deep, sau2016deep, kang2020towards}. 
For example, Sau et al.~\cite{sau2016deep} proposed an approach to simulate the effect of multiple teachers by injecting noise to the training data and perturbing the logit outputs of a teacher. In such a way, the perturbed outputs not only simulate the setting of multiple teachers but also result in noise in the softmax layer, thus regularizing the distillation loss. 
With the help of many teachers, the student is improved a lot. 
Kang et al.~\cite{kang2020towards} used Neural Architecture Search (NAS) to acquire knowledge for both the architecture and the parameters of the student network from different teachers. 
Besides the classic image classification task, KD can also be used in many other different fields, such as face recognition~\cite{ge2020efficient},  visual question answering~\cite{NIPS2018_8031}, video tasks~\cite{FuUltrafast, li2019spatiotemporal} etc. 

Recently, some KD methods have been proposed to have less or more than two stages. For one thing, KD can be a one-stage strategy~\cite{yang2019snapshot, zhang2018deep}. 
Zhang et al.~\cite{zhang2018deep} proposed that a pool of untrained student networks with the same network structure can be used to simultaneously learn the target task together instead of the traditional two-stage knowledge distillation strategy. 
For another, a line of research~\cite{mirzadeh2019improved, choi2020block} focuses on KD methods with more than two stages. 
In~\cite{mirzadeh2019improved}, several Teacher Assistant networks are used to transfer the knowledge from a teacher more softly and effectively. The teacher propagates its knowledge to the assistant networks first and then the assistant networks propagate its knowledge to the student network. 
In~\cite{choi2020block}, the student network has the same architecture as the teacher network at the beginning. 
The last block of the student network is replaced by a simple block and trained at the first stage. 
Next, the penultimate block is replaced similarly and the last two blocks are trained. 
In this style, all blocks are trained after several stages and a lightweight student network is achieved in the end. 

In this paper, we mainly focus on how to use the gap between the teacher and the student as knowledge. 
We use a series of lightweight networks, named res-students, to learn the gap in a multi-stage manner. 

\subsection{Ensemble Methods} 
Ensemble methods, which have been studied extensively for improving model performance \cite{hansen1990neural, dietterich2000ensemble, opitz1999popular}, are strategies to combine models by averaging, majority voting or something else, which means several models having the same status are used to improving final performance. 
Ensembles of models perform at least as well as each of its ensemble members~\cite{krogh1995neural}. 
There are several lines of research of ensemble methods: introducing different regularization, reducing training time~\cite{xie2013horizontal} and saving test time~\cite{bucilua2006model}.
 
For combining knowledge distillation and the ensemble idea, Lan et al.~\cite{lan2018knowledge} proposed to aggregate the logits of several homogeneous student models to become an ensemble teacher and then to distill the knowledge from the teacher. 
On the contrary, we first carry on knowledge distillation to build a student and res-students and then combine them. Also, the roles that the student and res-students play are not the same. The student acquires the knowledge from the teacher, while the res-students acquire the knowledge from knowledge gaps. Furthermore, we introduce a sample-adaptive strategy to decide which res-students to use at inference time.


\subsection{Residual Learning}
Residual learning has been studied in face alignment for improving the final performance by learning from the gap between the suboptimal results and the target.
Specifically, regressors for face alignment refine the current suboptimal results (landmark \mbox{\cite{park2020acn,cao2014face,ren2014face,zhu2016face}}, shape estimation~\cite{yang2017stacked} or heatmaps~\cite{wan2020robust}) to approach the target. 
In~\cite{ren2014face}, local binary features are extracted first and then a learning global linear regression is applied to generate the final landmark. 
In~\cite{yang2017stacked}, a cascade shape regression are employed to generate the optimal pose from the initial pose according to the shape-indexed features of the input image. 
In~\cite{wan2020robust}, heatmaps and suboptimal landmarks are extracted by a multi-order cross geometry-aware model from the original images and a multi-order high-precision hourglass network is applied to achieve the heatmap subpixel face alignment and generate the optimal landmarks.


Knowledge distillation aims to deliver a lightweight student network to approximate a given teacher network. At training time, our ResKD learns not only from the ground truth but also from the residual between the teacher network and the former student network in a coarse-to-fine manner. 
When learning from the teacher network, our residual-guided supervision can be applied at different levels: the final output (just as the ground truth), intermediate feature maps, structural knowledge, etc.

%% file: sections/method.tex
\section{Residual-Guided Knowledge Distillation}
\label{sec:res-kd}

\begin{figure*}[tb]
	\centering
	\includegraphics[width=\textwidth]{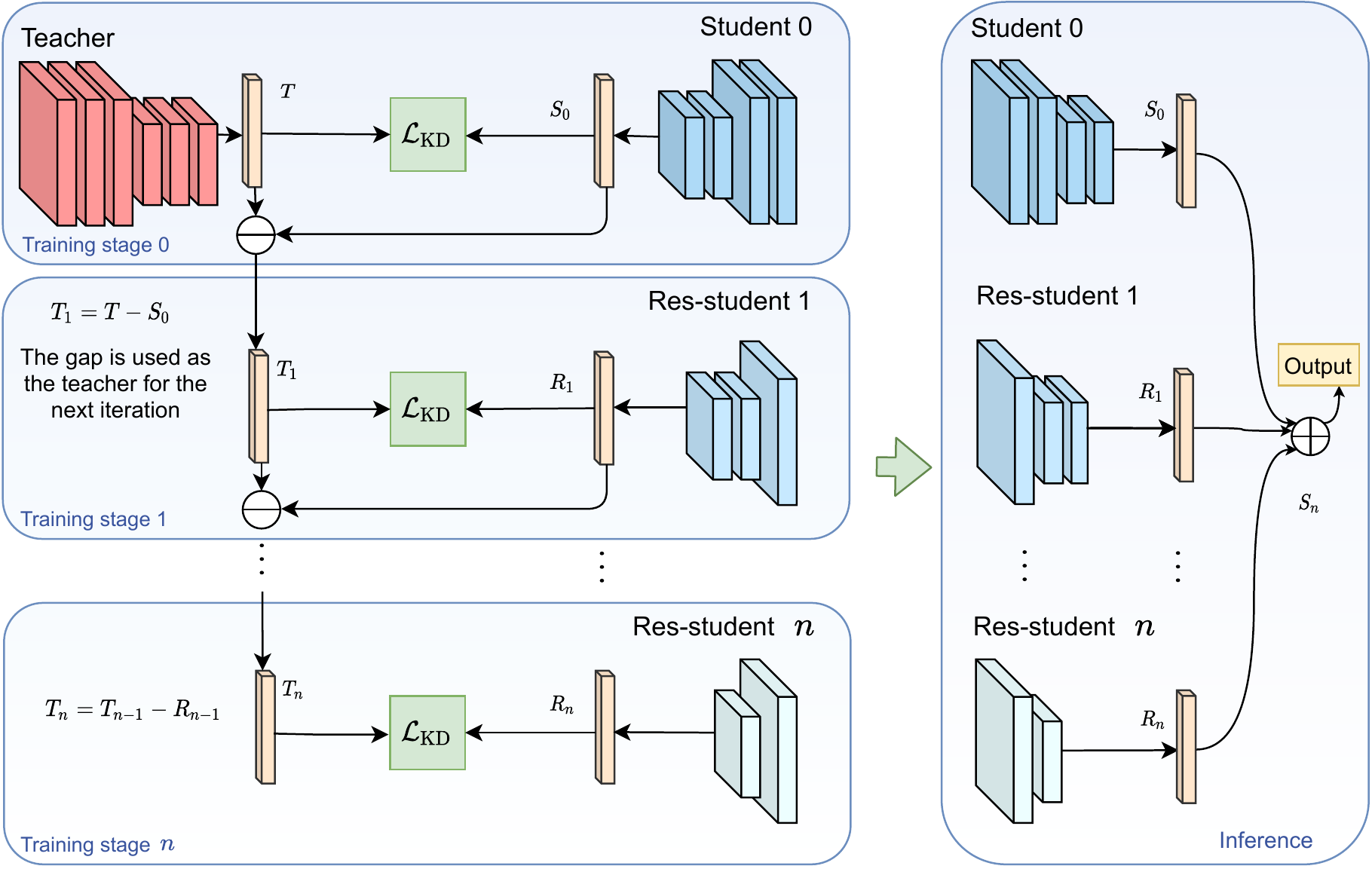}
	\caption{ (Best viewed in color) The main idea of the ResKD framework. Training stage $0$ is just the same as traditional KD: a student network $S_0$ is trained to learn from a teacher $T$. In stage $1$, a res-student $R_1$ is trained to learn from a teacher $T_1$ which is the residual between $T$ and $S_0$. Similarly, in stage $i$, a res-student $R_i$ is trained from a teacher $T_i$ until $R_n$ is obtained. $S_0$ and all the res-students are combined to build a new student $S_n$.} \label{fig:framework}
\end{figure*}

In this section, we present our residual-guided knowledge distillation framework (ResKD).
First, we briefly introduce the background. 
Second, we describe our main idea of residual-guided learning. 
Next, we propose a sample-adaptive strategy at inference time.
Finally, we put all the things together to build a whole framework. 
For convenience, \cref{tab:notation} summarizes the notations we use in this section.

\input{tables/notation}

\subsection{Background and Notations}
We first describe a general formulation of knowledge distillation, and then introduce the residual-guided knowledge distillation in this formulation. 

We define a teacher network $T$ as a function 
\begin{equation}
	T=f(\mathbf{x}, w_T, \alpha_T),
\end{equation}
where $\mathbf{x}$ is an input image, w$_{T}$ is the weights of $T$, and $\alpha$$_{T}$ is its architecture parameters. 
Similarly, $S=f(\mathbf{x}, w_{S}, \alpha_{S})$ denotes a student network. 
For convenience, a network and its logits are used interchangeably.  
The goal of knowledge distillation is to learn $w_{S}$ so that the results of the student $S$ are as close as possible to the teacher $T$.

Classic knowledge distillation from a teacher is done in two steps. 
First, the student architecture $\alpha_S$ is determined. 
Then, $w_{S}$ is optimized by
\begin{equation}
\label{eq:KD_ARG}
\operatorname*{argmin}\limits_{w_S}{\sum_{j=1}^{N}\mathcal{L}_{\mathrm{KD}}\left(f(\mathbf{x}^{(j)}, w_{S}, \alpha_{S}) ,f(\mathbf{x}^{(j)}, w_{T}, \alpha_{T})\right)},
\end{equation}
where $\mathbf{x}^{(j)}$ is an input image in a dataset with $N$ training samples and $\mathcal{L}_{\mathrm{KD}}$ is defined as
\begin{equation}
\begin{split}
\label{eq:KD_LOSS}
\mathcal{L}_{\mathrm{KD}}(S, y^{(j)} , T, \tau, t) =	\tau \cdot t^{2} \cdot  \mathcal{L}_{\mathrm{T-S}}(\sigma(\frac{S}{t}), \sigma(\frac{T}{t})) \\
	+ (1-\tau) \cdot \mathcal{L}_{\mathrm{CE}}(\sigma(S), y^{(j)}),
\end{split}
\end{equation}
where $y^{(j)}$ is the label of the input image $\mathbf{x}^{(j)}$, $\sigma(\cdot)$ is the softmax function, $\mathcal{L}_{\mathrm{CE}}$ is the conventional cross-entropy loss function, $\tau$ and $t$ are scalar hyper-parameters, and  $\mathcal{L}_{\mathrm{T-S}}$ is the loss between the student's logits and the teacher's logits, e.g., the Kullback--Leibler divergence.

The optimized $w_{S}$ from \cref{eq:KD_ARG} makes the predictions obtained from $S = f(\mathbf{x}, w_{S}, \alpha_{S})$ as close to those obtained by $T = f(\mathbf{x}, w_{T}, \alpha_{T})$ as possible, but we still observe a difference $\Delta$ between the teacher's and student's logits:
\begin{equation}
\label{eq:Delta}
\Delta = T - S,
\end{equation}
which is what we propose to reduce as follows.

\subsection{Residual-Guided Learning}
\label{ssub:res-learning}


We propose a framework distilling knowledge in a residual-guided fashion: we use the difference, or the \emph{residual}, between a teacher and a student as guidance. 
Given a teacher network, we first train a student network by classic knowledge distillation.  
Then, we introduce another student, called a res-student, to learn their residual. 
After that, we combine the original student and the res-student to form a new student. 
Such a process can be repeated to add more res-students to further bridge the performance gap, as shown in \cref{fig:framework}.


Formally, let the logits of the teacher be $T$, the logits of the first student be $S_0$, and the logits of the res-student networks be $\{R_i\}_{i=1}^n$. 
We train $S_0$ and $\{R_i\}_{i=1}^n$ in stages. In stage $0$, $S_0$ is trained in a classic knowledge distillation manner with $T$ as the teacher network. The gap can be observed by $\Delta_0 = T - S_0$. In stage $1$, the gap $\Delta_0$ is defined as a new teacher, i.e., $T_1 = \Delta_0$, to train a res-student $R_1$. 
Once we obtain $R_1$, we can define a new student $S_1 = S_0 + R_1$. Similarly, in each following stage, we use the gap from the current stage $\Delta_i = T_{i} - R_{i}$ as the next stage teacher network $T_{i+1}$. We distill a res-student $R_{i+1}$ from $T_{i+1}$ and obtain the next student $S_{i+1} = S_0 + R_1 + \cdots + R_{i+1}$.
Finally, we use the student network and res-student networks trained in all $n+1$ stages. 
We feed an input image to all networks independently and sum their logits to obtain the final logits:
\begin{equation}
\begin{split}
\label{eq:test_logits}
S_n = S_0 + \sum_{i=1}^{n} R_{i}.
\end{split}
\end{equation}


\subsubsection{NAS-Assisted Architecture} 
When we use a res-student to bridge the gap to the teacher, it is better to use NAS to get the architecture of the res-student network instead of a handcrafted one. 
We also use the knowledge distillation loss function in the search stage when using NAS, which is reflected by adding $\alpha_S$ as an optimizable parameter in \cref{eq:KD_ARG}:
\begin{equation}
\label{NASKD_ARG}
\operatorname*{argmin}\limits_{w_S,\alpha_S}{\sum_{j=1}^{N}\mathcal{L}_{\mathrm{KD}}\left(f(\mathbf{x}^{(j)}, w_S, \alpha_S) ,f(\mathbf{x}^{(j)}, w_{T}, \alpha_{T})\right)}.
\end{equation}

We use the same search space as that of STACNAS~\cite{li2019stacnas}. 
A STACNAS-style neural network consists of a series of building blocks called cells. 
We define a set of candidate operations $O$ to be used inside these cells, and from which we will select the best operations for the architecture.

\begin{figure}[!tb]
	\centering
	\includegraphics[width=\linewidth]{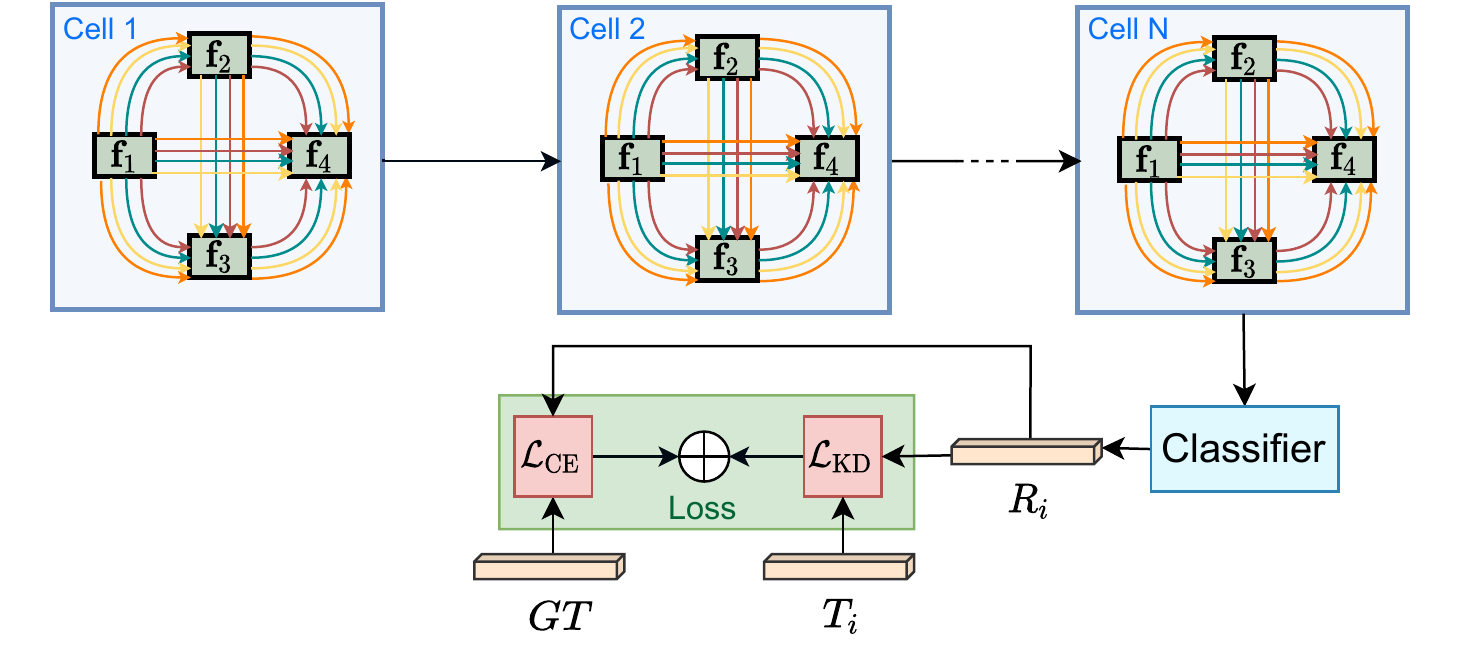}
	\caption{The overview of our NAS-assisted architecture. We use the same setting as that of STACNAS\cite{li2019stacnas}. In addition, we use KD loss in both search and fine-tuning stages.}
	\label{exp_2_example}
	\centering
\end{figure}

The construction of cells and the whole network can be described as a directed acyclic graph (DAG) as shown in \cref{exp_2_example}.
A cell contains a sequence of $N$ nodes ${\mathbf{f}_{1},\mathbf{f}_{2}, \cdots, \mathbf{f}_{N}}$ each of which is a stack of  feature maps. These nodes are connected by directed edges. An edge $(i,j)$ represents some operation $o_k^{(i,j)}(\cdot)\in O$ that transforms $\mathbf{f}_i$ to $\mathbf{f}_j$. 
In a certain cell, each node is computed as a weighted sum of all its predecessors:

\begin{equation}
\label{alphaO_compute}
\mathbf{f}_{j} = \sum_{i<j}\sum_{o_k\in O}\alpha_k^{(i,j)}o_k^{(i,j)}(\mathbf{f}_{i}),
\end{equation}
where $\alpha_k^{(i,j)}$ is the weight for the operation $o_k^{(i,j)}$ applied to $\mathbf{f}_i$ when calculating the node $\mathbf{f}_j$. We use the same training strategy as that of STACNAS~\cite{li2019stacnas} to obtain the final suitable network.

\subsubsection{Termination Condition}
When a ResKD student $S_i$ is close enough to the teacher $T$, we can stop the residual-guided process. Here, we define a concept ``Energy''  to measure the difference between a student and a teacher.

The Energy of network $S_i$ for a certain sample $\mathbf{x}^{(j)}$ is 
\begin{equation}
\begin{split}
\label{eq:Delta2}
\mathrm{Energy}(S_i, \{\mathbf{x}^{(j)}\}) = \|\sigma(S_i(\mathbf{x}^{(j)}))\|_2^2,
\end{split}
\end{equation}
where $\sigma(\cdot)$ is the softmax function, and thus the Energy of network $S_i$ on dataset $D$ is
\begin{equation}
\begin{split}
\label{eq:Delta2}
\mathrm{Energy}(S_i, D) = \frac{1}{N} \cdot \sum_{j=1}^{N} \|\sigma(S_i(\mathbf{x}^{(j)}))\|_2^2,
\end{split}
\end{equation}
where $N$ is the total number of samples in dataset $D$.
In this way, the Energy of a network represents its overall confidence on a dataset.
When the Energy of $S_{i}$ has reached a comparable level of the Teacher's Energy (e.g., $90\%$), we can set $n = i$ and finish the residual-guided learning process.

In practice, we calculate Energy on a validation set $D_v$ whose data are uniformly sampled from the training set $D$. The overall training process is shown in \cref{algo_training}.


\begin{algorithm}[tb]
	\caption{ResKD training} 
	\label{algo_training}
	\begin{algorithmic}
	\REQUIRE Training set $D_t = \{(\mathbf{x}^{(j)}, y^{(j)})\}_{j=1}^{N}$ and teacher network $T(\mathbf{x}) = f(\mathbf{x}, w_T, \alpha_T)$.
	\ENSURE $S_0, R_1, R_2, \cdots, R_n$ and a threshold $\mathrm{TH}_{\mathrm{energy}}$.
	\STATE Sample a validation set $D_v = \{(\mathbf{x}^{(j)}, y^{(j)})\}_{j=1}^{N_v}$ from the training set $D_t$ uniformly.
	\STATE Train $S_0$; $T_1 = \Delta_{0} = T - S_0$; $i = 0$.\\
	\REPEAT
	\STATE $i = i + 1$.
	\STATE $R_i = \mathrm{KD}(T_i)$. \\
	\STATE $T_{i+1} = T_i - R_i$; $S_{i} = S_{i-1} + R_i$. \\
	\UNTIL{$\mathrm{Energy}(S_i, D_v)>90\% \cdot \mathrm{Energy}(T, D_v)$}
	\STATE $n = i$.
	\STATE $\mathrm{TH}_{\mathrm{energy}} = \mathrm{Energy}(S_n, D_v)$
	\RETURN networks $S_0, R_1, \cdots, R_n$ and a threshold $\mathrm{TH}_{\mathrm{energy}}$.
	\end{algorithmic}
	\end{algorithm}

\subsection{Sample-Adaptive Inference}
\label{ssec:sample-adaptive-inference}

When we finish the residual-guided learning process, we have $S_n = S_0 + \sum_{i=1}^{n} R_{i}$, where $R_i$ is supposed to bridge the gap $\Delta_{i-1}$ between $S_{i-1}$ and $T$. 
However, for each sample, $\Delta_{i-1}$ is also different. For instance, if the sample is easy to recognize, $\Delta_0$ can be subtle, and if the sample is difficult, even $\Delta_2$ can be considerable. In other words, for an easy sample, $S_0$ is enough, but for a difficult sample, we should use more res-students. 

Based on the above observation, we propose an adaptive strategy for each sample at inference time. 
Similar to the Energy idea in \cref{ssub:res-learning}, we uniformly sample a validation set $D_v$ from the training set $D$, and calculate an Energy threshold for the final ResKD student $S_n$:
\begin{equation}
\label{eq:th_energy}
	\begin{split}
		\mathrm{TH}_{\mathrm{energy}} = \mathrm{Energy}(S_n, D_v).
	\end{split}
\end{equation}
When the Energy of $S_i$ with a sample $\mathbf{x}^{(j)}$, i.e., $\mathrm{Energy}(S_i, \{\mathbf{x}^{(j)}\})$, is higher than $\mathrm{TH}_{\mathrm{energy}}$, we set the rest res-students $R_{i+1}, \cdots, R_n$ aside.
As a result, given a sample $\mathbf{x}^{(j)}$, we define $S_n^{(i)}$ as the student network it uses at inference time:
\begin{equation}
\begin{split}
\label{ensemble}
S_n^{(i)} (\mathbf{x}^{(j)}) = S_L (\mathbf{x}^{(j)}), \ 0 \leq L \leq n,
\end{split}
\end{equation}
where we use $S_L$ instead of $S_n$ for the sample $\mathbf{x}^{(j)}$. For example, let $S_n = S_3$. Given a sample $\mathbf{x}^{(j)}$, if $\mathrm{Energy}(S_1, \mathbf{x}^{(j)}) > \mathrm{TH_{energy}}$, then $L=1$, which means that we only use $S_0$ and $R_1$, and set $R_2$ and $R_3$ aside. The whole process is shown in \cref{algo_inference}.

\subsection{ResKD: The Whole Framework}
When faced with a knowledge distillation problem, our first step is to train a student network $S_0$, which can be handcrafted or searched, under the guidance of the teacher $T$.
Next, we start to use our residual-guided learning strategy to find res-students. 
We train the res-student $R_i$  in a KD manner under the guidance of $T_i$.
Such a residual-guided process can be repeated until $S_i$ has achieved the Energy comparable to $T$. At inference time, we apply our sample-adaptive strategy.  


\begin{algorithm}[tb]
	\caption{Sample-adaptive inference} 
	\label{algo_inference}
	\begin{algorithmic}
	\REQUIRE a test sample $\mathbf{x}^{(j)}$, networks $S_0, R_1, R_2, \cdots, R_n$ and a threshold $\mathrm{TH}_{\mathrm{energy}}$.
	\ENSURE $S_L$ and the logits of $S_L$ for the test sample $\mathbf{x}^{(j)}$.
	\STATE calculate $S_0 (\mathbf{x}^{(j)})$; $E_0 = \mathrm{Energy}(S_0, \{\mathbf{x}^{(j)}\})$.
	\STATE $i = 0$.
	\WHILE {$\mathrm{TH}_{\mathrm{energy}} \ge E_i$ and $i < n$.}
	\STATE calculate $R_i(\mathbf{x}^{(j)})$; $i = i +1$\\
	\STATE $E_i = E_{i-1} + \mathrm{Energy}(R_i, \{\mathbf{x}^{(j)})\}$.\\
	\ENDWHILE
	\STATE $L = i$.
	\RETURN $S_L$ and the logits of $S_L$ for the test sample $\mathbf{x}^{(j)}$.
	\end{algorithmic}
\end{algorithm}

%% file: tables/notation.tex
\begin{table}[tb]
\begin{center}
	\caption{Notations}
\label{tab:notation}
\begin{tabular}{ll}
\toprule
$T$ & A teacher network or its logits \\
$S_0$ & A classic student network or its logits \\
$S_i, i>0$ & A ResKD student network or its logits \\
$R_i$ & A res-student network or its logits \\
$\Delta_{0}$ & The gap between $S_{0}$ and $T$ \\
$\Delta_{i}, i>0$ & The gap between $S_i$ and $T_i$ \\
$D_t$ & A training set \\
$D_v$ & A validation set \\
$\mathcal{L}_\mathrm{KD}$ & The knowledge distillation loss function \\
$\mathcal{L}_{\mathrm{T-S}}$ & The loss function used between logits of $T$ and $S$ \\
$\sigma(\cdot)$ & The Softmax function \\
$\mathbf{f_j}$ & Feature maps \\

\bottomrule
\end{tabular}
\end{center}
\end{table}

%% file: sections/experiment.tex
\section{Experiments}

In this section, we first introduce the datasets and protocols we use in \cref{ssec:datasets}. 
Next, we do ablation studies in \cref{ssec:Residualguided} and \cref{ssec:Sampleadaptive} to evaluate the effectiveness of our strategies.
Finally, we evaluate our framework on several datasets in \cref{ssec:more}. 

\subsection{Datasets and Protocols}
\label{ssec:datasets}

\subsubsection{CIFAR-10/100}
The CIFAR-10/100 dataset consists of 50k training images and 10k testing images in 10/100 classes.
We use a weight decay of 0.0001 and a momentum of 0.9. 
Our models are trained with a mini-batch size of 128. 
We start with a learning rate of 0.1, divide it by 10 at 150 and 200 epochs, and terminate training at 250 epochs. 
We follow the simple data augmentation in~\cite{he2016deep} for training: 4 pixels are padded on each side, and a $32 \times 32$ crop is randomly sampled from the padded image or its horizontal flip for training. 
For testing, we only evaluate the single view of the original $32 \times 32$ image.

\subsubsection{ImageNet}
ImageNet consists of 1000 classes. 
Our models are trained on the 1.28 million training images, and evaluated on the 50k validation images.
Images are randomly resized and a $224 \times 224$ crop is randomly sampled from the resized image or its horizontal flip for training. 
For testing, we scale the short side of images to 224, and a $224 \times 224$ center crop is sampled from the scaled image.
When training ResNet series networks, We start with a learning rate of 0.1, divide it by 10 at 30, 60 and 80 epochs, and terminate training at 90 epochs. 
We use a weight decay of 0.0001 and a momentum of 0.9. 
When dealing with the network given by neural architecture search, we use a weight decay of 0.0003 and a momentum of 0.9. 
An exponential learning rate strategy is used to control the learning rate from 0.1 and multiplying the learning rate by 0.92 after each epoch. We use 100 epochs in total.
Our models are trained with a mini-batch size of 128. 

\subsubsection{Tiny-ImageNet}
Tiny-ImageNet consists of 200 classes. 
Our models are trained on the 100k training images, and evaluated on the 10k validation images.
Images are randomly resized and a $224 \times 224$ crop is randomly sampled from the resized image or its horizontal flip for training. 
For testing, we resize the original images to $224 \times 224$ and evaluate them.
When training ResNet series networks, we start with a learning rate of 0.1, divide it by 10 at 40, 80, and 120 epochs, and terminate training at 150 epochs. 
We use a weight decay of 0.0001 and momentum of 0.9. 
When dealing with the network given by neural architecture search, we use the same setting as we do on ImageNet. 
Our models are trained with a mini-batch size of 128.

When it comes to the termination condition of residual-guided learning, we set comparable Energy as 90$\%$ of $T$'s Energy. 
In the following sections, when it is not mentioned, we use $L_2$ loss function for the $\mathcal{L}_{\mathrm{T-S}}$ in \cref{eq:KD_LOSS}. ``KD'' in some tables (e.g., \cref{tab:L2_KL}) means using this loss function, too.

We do all experiments on four NVIDIA GTX 1080 Ti  cards.

\input{tables/gradual_comp}



\input{tables/num_comp}

\subsection{Ablation Study}

\subsubsection{Residual-Guided Learning}
\label{ssec:Residualguided}

First, we validate the effect of a res-student. We use ResNet-110 as the teacher $T$, ResNet-20 as the classic KD student $S_0$, and ResNet-14 as the res-student $R_1$. As illustrated in \cref{tab:comp_ensemble}, our ResKD student $S_1 = S_0 + R_1$  outperforms $S_0$, which means a res-student can correct the errors of a classic KD student. 

However, since we obtain a ResKD student by combining two networks, it is arguable whether the effect of the res-student comes from our residual-guided learning. It could be attributed to additional capacity. Thus, it is necessary to compare a ResKD student with other types of combinations, such as an ensemble network, to show the effect of our method. As illustrated in \cref{tab:comp_ensemble}, our method also outperforms the ``ensemble'' and the ``trained together'' approaches, which means that using res-students is more effective than using an ensemble.

We can repeat residual-guided learning to further bridge the gap between the student and the teacher.
As illustrated in \cref{num_comp}, we train another res-student $R_2$ to build $S_2$, and the performance of $S_2$ is better than $S_1$.

\subsubsection{Sample-Adaptive Inference} \label{ssec:Sampleadaptive}
This strategy focuses on striking the balance between performance and cost. 
Our threshold $\mathrm{TH}_{\mathrm{energy}}$ based on the validation set is a key hyper-parameter in it. 
We justify our choice of $\mathrm{TH}_{\mathrm{energy}}$ in two fields: our $\mathrm{TH}_{\mathrm{energy}}$ can be generalized to the test set, and the $\mathrm{TH}_{\mathrm{energy}}$ is effective. 

\input{tables/10_100_final}

\input{tables/tiny_image_final}

\begin{figure}[tb]
    \begin{subfigure}{0.45\linewidth}
        \centering
        \includegraphics[width=1\linewidth]{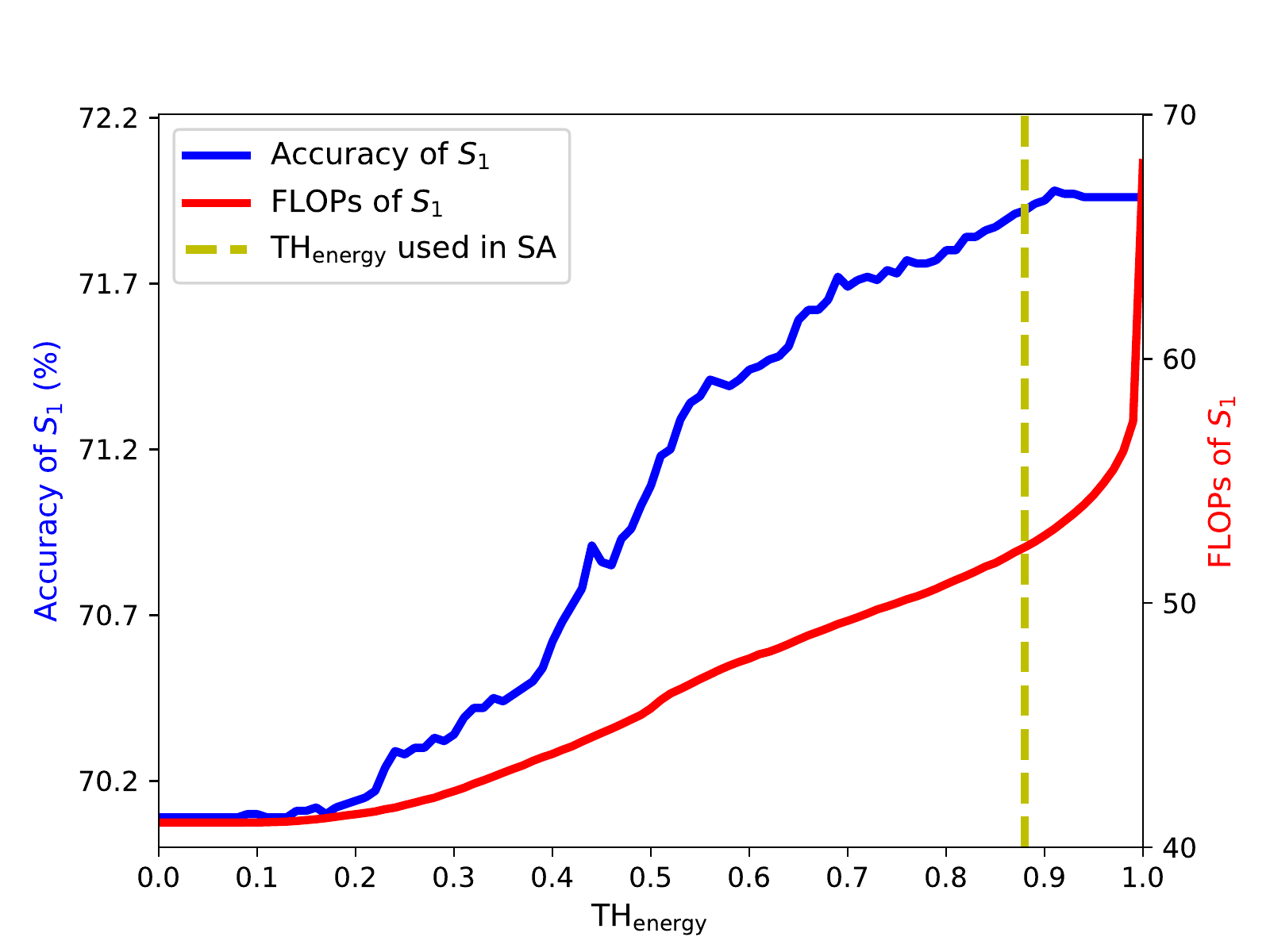}\\
        \caption{$S_1$'s accuracy and FLOPs with different $\mathrm{TH}_{\mathrm{energy}}$ on CIFAR-100.} \label{TH100}
    \end{subfigure}%
    \hspace{1em}
    \begin{subfigure}{0.45\linewidth}
        \centering
        \includegraphics[width=1\linewidth]{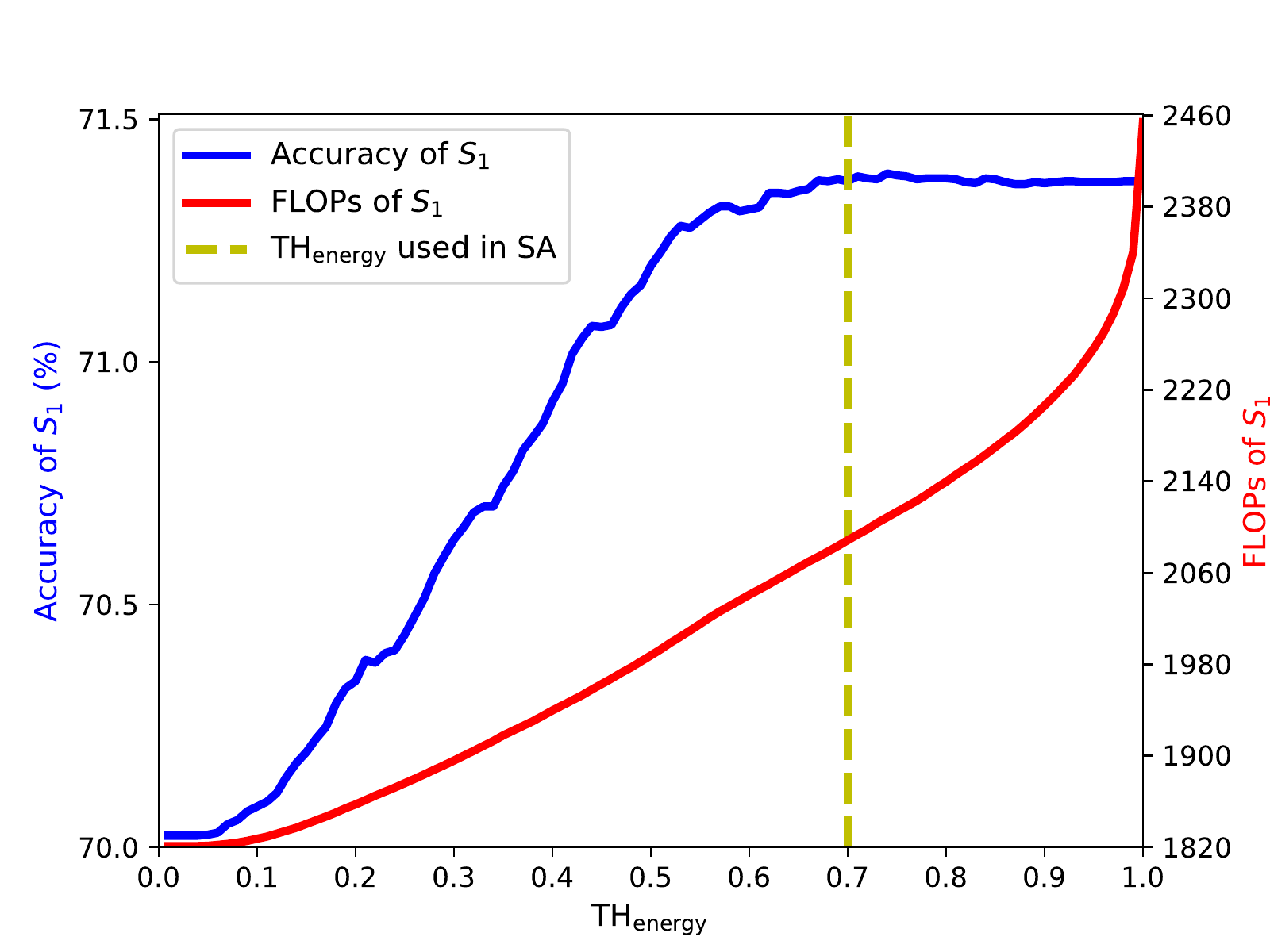}\\
        \caption{$S_1$'s accuracy and FLOPs with different $\mathrm{TH}_{\mathrm{energy}}$ on ImageNet.} \label{THimage}
    \end{subfigure}%

    \centering
    \caption{This figure shows the results about $\mathrm{TH}_{\mathrm{energy}}$.
    (a) indicates the relationship between the performance and FLOPs of $S_1$ with different $\mathrm{TH}_{\mathrm{energy}}$ on the test set on CIFAR-100. 
    (b) shows the relationship on ImageNet. 
    For CIFAR-100, $T$ is ResNet-110, $S_0$ is ResNet-20, and $R_{1}$ is ResNet-14. For ImageNet, $T$ is ResNet-50, $S_0$ is ResNet-18, and $R_{1}$ is a NAS-based architecture. The results indicate that our trade-off is done by our sample-adaptive strategy, and most of $R_1$'s FLOPs can be reduced with little performance decreasing.
    } \label{TH2pict}
\end{figure}

Our sample-adaptive strategy focuses on striking the balance between performance and cost by setting $\mathrm{TH}_{\mathrm{energy}}$. 
According to the relationship between performance and cost w.r.t. $\mathrm{TH}_{\mathrm{energy}}$ as shown in \cref{TH2pict}, we can set an appropriate $\mathrm{TH}_{\mathrm{energy}}$.
We set $\mathrm{TH}_{\mathrm{energy}}$ as 90\% of the final student network's energy (here is $S_1$'s energy) on the validation set to have a good trade-off. As a result, $\mathrm{TH}_{\mathrm{energy}}$ is 0.88 for CIFAR-100 and 0.70 for ImageNet. In this case, our sample-adaptive strategy can avoid remarkable FLOPs of $R_1$ and preserve the accuracy at the same time.

\begin{figure}[tb]
	\centering
	\includegraphics[width=0.6\linewidth]{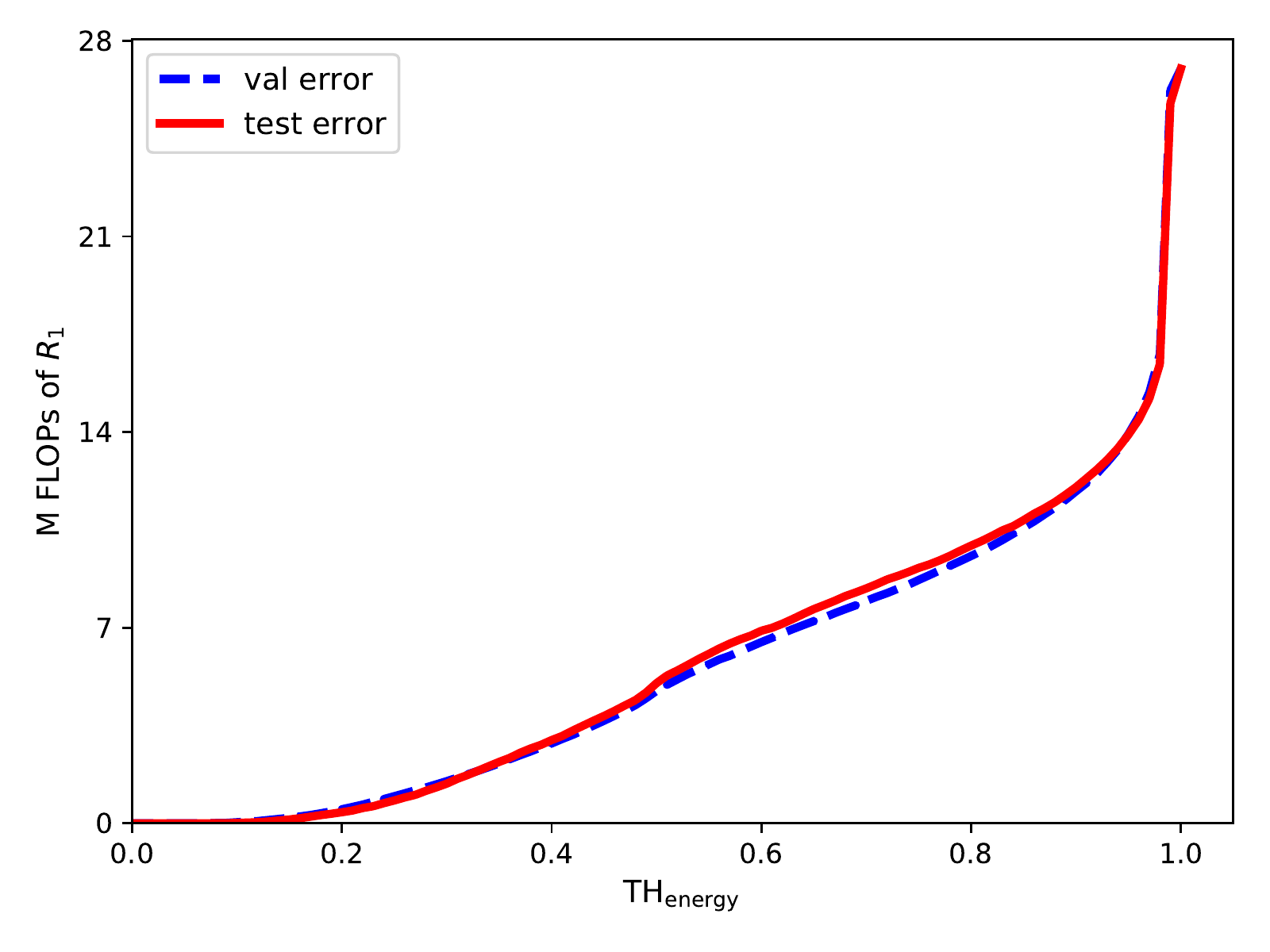}
	\caption{The relationship between the cost reduction of $R_1$ and $\mathrm{TH}_{\mathrm{energy}}$ on  both validation set and test set on CIFAR-100. $T$ is ResNet-110, $S_0$ is ResNet-20, and $R_{1}$ is ResNet-14. The result indicates that it is reasonable to calculate  $\mathrm{TH}_{\mathrm{energy}}$ on a validation set for the sample-adaptive strategy.}
	\label{THfig}
\end{figure}

We also verify the transfer from our validation set to the test set. We carry out experiments to show the relationship between $\mathrm{TH}_{\mathrm{energy}}$ and cost on the validation and test set in \cref{THfig}. 
We can learn that as the $\mathrm{TH}_{\mathrm{energy}}$ reduces, the decreasing styles of $R_1$'s cost on both settings are similar, which means the $\mathrm{TH}_{\mathrm{energy}}$ based on the validation set is reasonable to use on the test set. 

\subsubsection{The Whole Framework, ResKD} 
\label{ssec:more}
We use our whole framework on CIFAR-10, CIFAR-100, Tiny-ImageNet and ImageNet.
As illustrated in \cref{tab:10_100_final} and \cref{tab:tiny_image_final}, ResKD achieves consistent results on these datasets. Our ResKD student $S_1 = S_0 + R_1$ where $R_1$'s architecture is searched by NAS outperforms $S_0$ in both accuracy and energy, which means a res-student can also correct the errors of a classic KD student on all four datasets.

\subsection{Discussion}

To verify the generalization of our ResKD, we carry out the following four case studies.

\input{tables/L2_KL}
\subsubsection{Different Loss Functions} To show that our residual-guided learning is generic to different $\mathcal{L}_{\mathrm{T-S}}$, we carry on experiments using Kullback--Leibler divergence and $L_2$ distance loss function on CIFAR-10. As illustrated in \cref{tab:L2_KL}, both Kullback--Leibler divergence  and $L_2$ distance loss function work well with our ResKD framework.

\subsubsection{Settings of $\tau$ and $t$} 
\begin{figure}[tb]
        \begin{subfigure}{0.5\linewidth}
            \centering
            \includegraphics[width=1\linewidth]{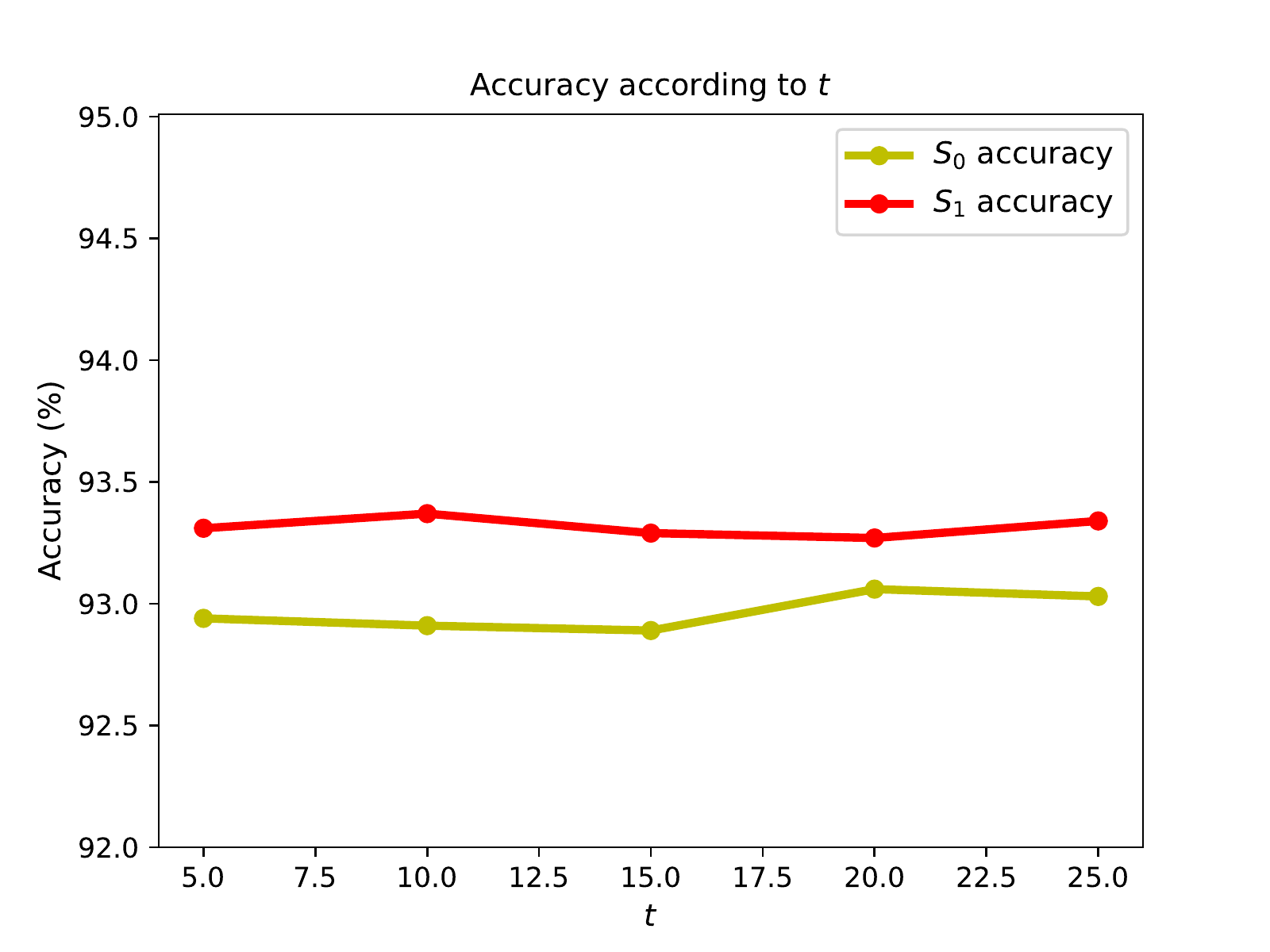}\\
            \caption{$t$'s influence on $S_0$ and $S_1$.} 
        \end{subfigure}%
        \begin{subfigure}{0.5\linewidth}
            \centering
            \includegraphics[width=1\linewidth]{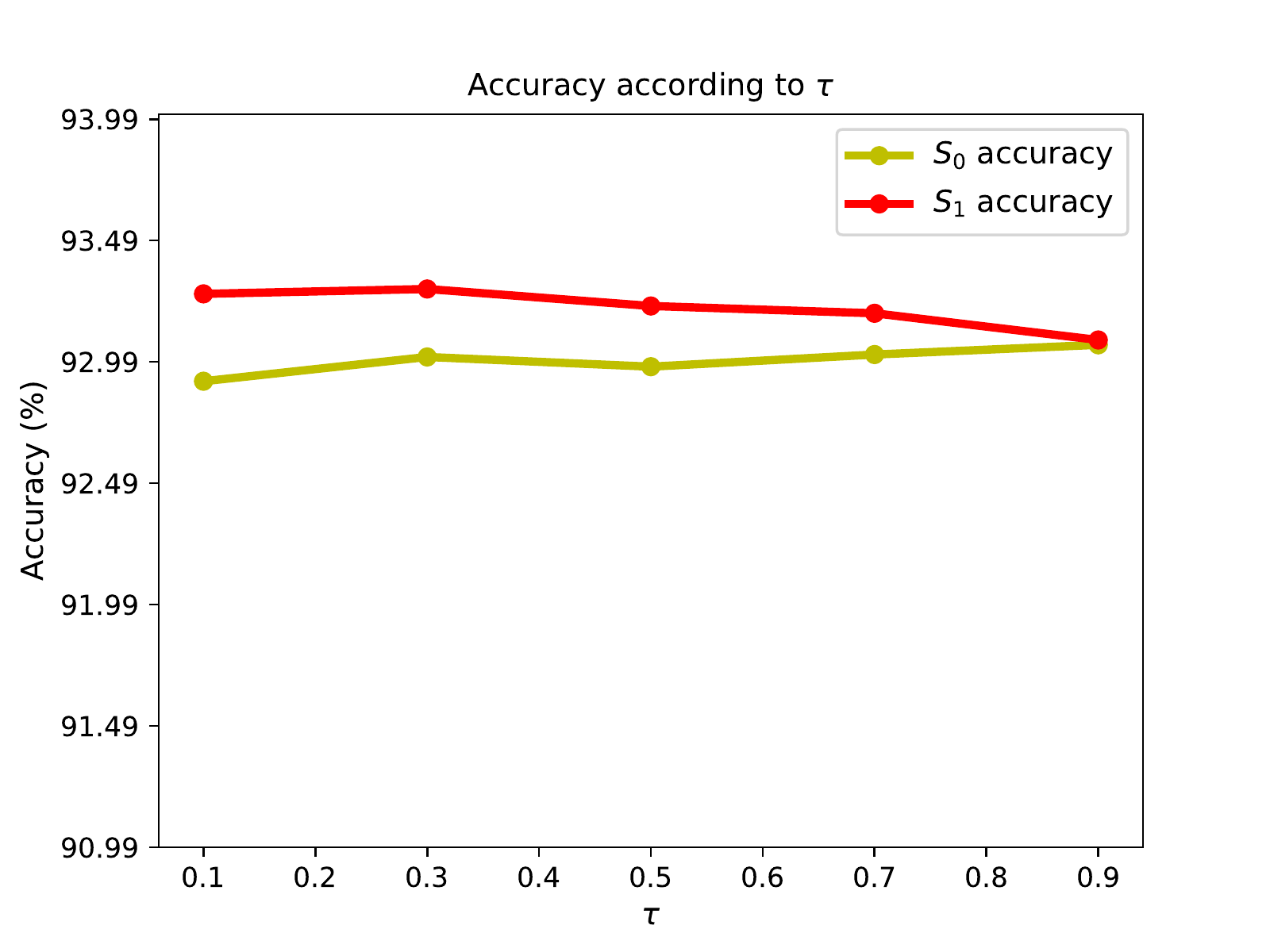}\\
            \caption{$\tau$'s influence on $S_0$ and $S_1$.} 
        \end{subfigure}%
    \centering
    \caption{The relationship between hyper-parameters ($\tau$ and $t$) and the performance of $S_0$ and $S_1$ on CIFAR-10. $T$ is ResNet-110, $S_0$ is ResNet-20, and $R_{1}$ is ResNet-14 The result indicates that these hyper-parameters can partly affect the final performance but not large.}
    \label{ttau}
\end{figure}

We have tried different $\tau$s (from 0.1 to 0.9) and $t$s  (from 5 to 25) and choose the best one. It turns out that the performance of ResKD is not very sensitive to these hyper-parameter configurations as shown in \cref{ttau}.
We set $\tau = 0.9 / 0.5 / 0.5 / 0.1$ for $S_0$ on CIFAR-10 / CIFAR-100 / Tiny-ImageNet / ImageNet, $\tau = 0.1$ for $R_1$ on all datasets and $t = 20$ in all situations.

\subsubsection{Different Backbones} To show that our framework is generic to different backbones, we also apply our residual-guided learning to other networks like VGG-11, MobileNetv2, ShuffleNetv2, and Xception. As shown in \cref{tab:backbone}, our framework achieves consistent results with different backbones.
\input{tables/backbone.tex}
\subsubsection{Accuracy-Cost Trade-off} 
\begin{figure}[h]
    \centering%
    \includegraphics[width=0.5\textwidth]{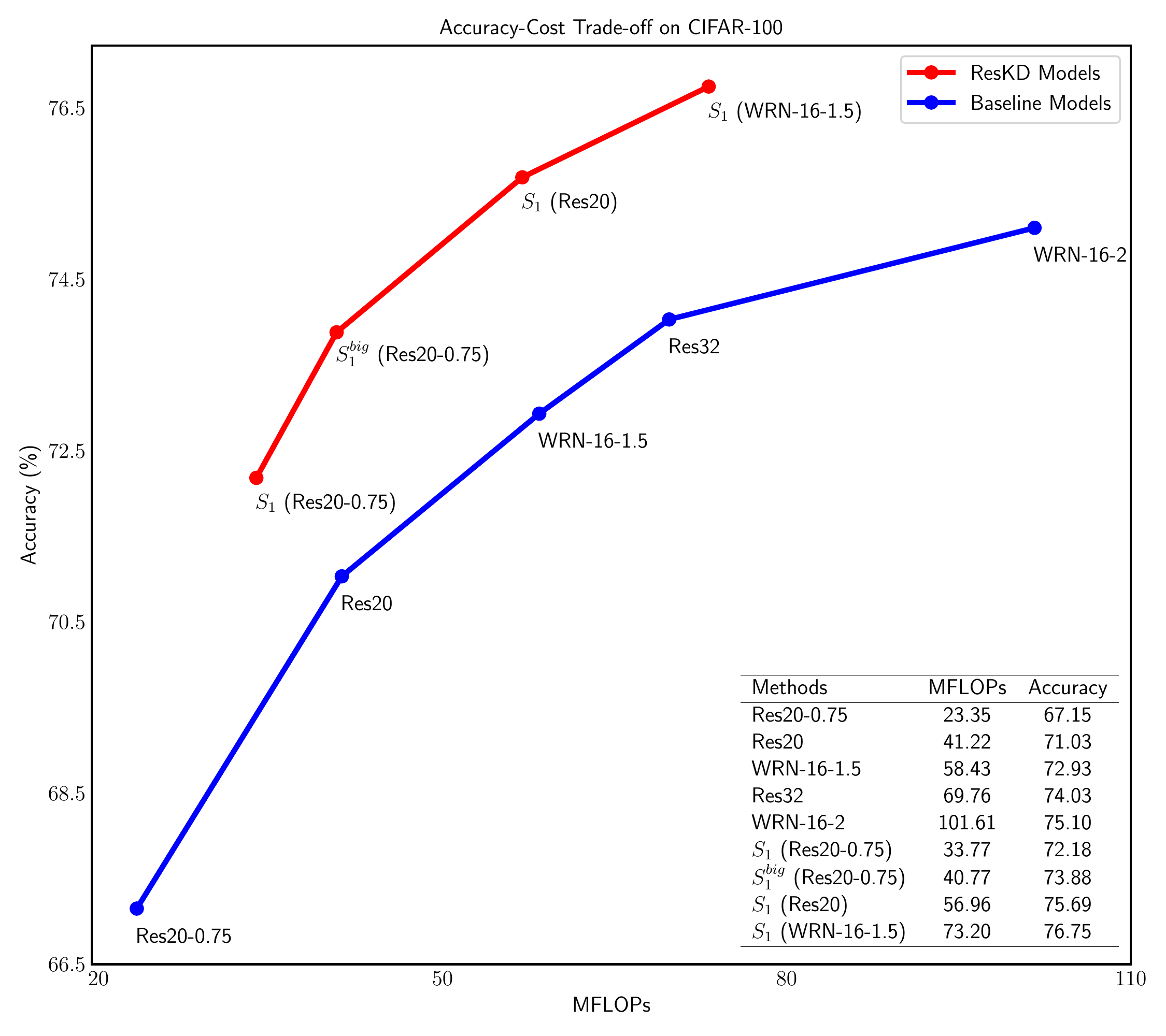}
    \caption{
    Accuracy-cost trade-off on CIFAR-100. 
    The blue line shows the accuracy-cost tendency of baseline models whereas the red line indicates the trend of ResKD models.
    ``$S_1$ (Res20)'': ResNet-20 is used as $S_0$;
    ``$S_1$ (Res20-0.75)'': the tailored ResNet-20 where the  channel number of each layer is 75\% of the original ResNet-20 is used as $S_0$;
    ``WRN-16-2'': the wide residual network whose depth and widening factor is 16 and 2;
     $S_1^{big}$ (Res20-0.75): similar to $S_1$ (Res20-0.75) but having a relatively big $R_1$.  
    We use the original knowledge distillation~\cite{hinton2015distilling} in this experiment. 
    } \label{fig.cost_acc}
\end{figure}

We show the comparison of the accuracy-cost trade-off between ResKD and the baseline as illustrated in \mbox{\cref{fig.cost_acc}}. With the same FLOPs, ResKD models achieve better accuracy, and when achieving the same accuracy, ResKD models demand less FLOPs.

\subsubsection{Different KD Methods}
    To show that our ResKD framework is generic to other distillation methods (feature distillation and both prediction and feature distillation) and for a fair comparison, we use the new implementation~\cite{tian2019contrastive}. 
    Experimental results show that our ResKD models achieve consistent accuracy improvement for all the twelve KD methods as shown in \mbox{\cref{table.KD_method20main}}.

\input{tables/KD_method20.tex}

%% file: tables/gradual_comp.tex
\begin{table}[tb]
\begin{center}
\caption{Comparison with other ensemble methods on CIFAR-10/100. Res110 is the ResNet-110 teacher network. Res20 (SGD) is a ResNet-20 trained with SGD, and Res20 (KD) is a ResNet-20 trained by KD to learn from the teacher. \emph{`Ensemble'}, \emph{`Trained together'} and \emph{`Ours'} share the same architecture where $S_0$ is ResNet-20 and $R_1$ is ResNet-14. 
`Ensemble' means that ResNet-20 and ResNet-14 are trained independently and the sum of their logits is used for evaluation.`Trained together' means that the combination of ResNet-20 and ResNet-14 is trained together.
}
\label{tab:comp_ensemble}
\begin{tabular}{lccc}
    \toprule
Architecture & Acc.~(\%) & \#Params (M) & MFLOPs\\
\midrule
\emph{Teacher}  & & &\\
Res110 & 94.19 / 72.44 & 1.728 / 1.734 & 255 / 255\\
\midrule
\emph{Students}  & & &\\
Res20 (SGD)  & 91.80 / 68.82 & 0.270 / 0.276 & 41 / 41\\
Res20 (KD)  & 93.06 / 70.09  & 0.270 / 0.276 & 41 / 41\\
\midrule
\emph{Res20-14}  & & &\\
ensemble & 93.17 / 70.41 & 0.442 / 0.454 & 68 / 68\\
trained together & 93.12 / 70.85  & 0.442 / 0.454 & 68 / 68\\
Ours & \textbf{93.27} / \textbf{71.96} & 0.442 / 0.454 & 68 / 68 \\
\bottomrule
\end{tabular}
\end{center}
\end{table}

%% file: tables/num_comp.tex

\begin{table}[tb]
\begin{center}
\caption{
Multi-stage residual-guided learning on CIFAR-10/100.    
Res$X$-$Y$-$Z$ means we use ResNet-$X$ as $S_{0}$, ResNet-$Y$ as $R_{1}$, and ResNet-$Z$ as $R_{2}$. 
Multi-stage residual-guided learning can further bridge the gap between the teacher and the student. \textbf{Bold}: The best results out of students. \ul{Underline}: The second best  results out of students.
}
\label{num_comp}
\begin{tabular}{lccc}
    \toprule
Architecture & Acc.~(\%) & \#Params (M) & MFLOPs\\
\midrule
\emph{Teacher}  & & &\\
Res110 & 94.19 / 72.44 & 1.728 / 1.734 & 255 / 255\\
\midrule
\emph{Students}  & & &\\
Res20 (SGD) & 91.80 / 68.82 & 0.270 / 0.276 & \textbf{41} / \textbf{41}\\
Res20 (KD)  & 93.06 / 70.09 & 0.270 / 0.276 & \textbf{41} / \textbf{41}\\
\midrule
\emph{ResKD $S_1$}  & & &\\

Res20-14 & 93.27 / 71.96 & 0.442 / 0.454 & \ul{68} / \ul{68}\\
Res20-20 & \ul{93.35} / \ul{72.22} & 0.539 / 0.551 & 82 / 82\\
\midrule
\emph{ResKD $S_2$}  & & &\\
Res20-14-8 &  93.30 /  72.18 & 0.518 / 0.535 & 80 / 80 \\
Res20-14-14 &  \textbf{93.68} /  \textbf{72.42} & 0.615 / 0.632 & 94 / 94 \\

\bottomrule
\end{tabular}
\end{center}
\end{table}

%% file: tables/10_100_final.tex
\begin{table*}[tb]
\begin{center}
\caption{
    Performance on CIFAR-10 / CIFAR-100. $T$ is ResNet-110, $S_0$ is a ResNet-20, $R_{1}$ is a res-student whose architecture is searched by NAS. $S_1$ means using both $S_0$ and $R_1$. `SA' means using our sample-adaptive strategy at inference time. Energy is the metric we mention in \cref{ssub:res-learning}.
     \textbf{Bold}: The best results out of students. \ul{Underline}: The second best  results out of students.
    }
\label{tab:10_100_final}
\resizebox{1\textwidth}{!}{
\begin{tabular}{lccccccc}
    \toprule
Architecture & Optimizer & SA & Acc.~(\%) & \#Params (M) & MFLOPs & MFLOPs proportion to $T$ (\%) & Energy \\
\midrule
\emph{Teacher}  & & & & &\\
Res110 & SGD &  & 94.19 / 72.44 & 1.728 / 1.734 &  255 / 255 & 100 / 100 & 0.9993 / 0.9894\\
\midrule
\emph{Students}  & & & & &\\
$S_0$ & KD &  & \ul{93.06} / \ul{70.09}  & 0.270 / 0.276 &  41 / 41 & \textbf{16.08} / \textbf{16.08}  &  0.9923 / 0.8380\\

\midrule
\emph{ResKD students}  & & & & &\\

$S_1$  & KD &  & \textbf{93.92} / \textbf{74.06} & 0.453 /  0.485 &  79 / 81 &  30.98 / 31.76 & 0.9958 / 0.9204\\
$S_1$  & KD & $\checkmark$ & \textbf{93.92} / \textbf{74.06} &  0.453 / 0.485 &  46 / 59 & \ul{18.04} / \ul{23.14} & -\\

\bottomrule
\end{tabular}
}
\end{center}
\end{table*}

%% file: tables/tiny_image_final.tex
\begin{table*}[tb]
    \begin{center}
    \caption{
        Performance on Tiny-ImageNet / ImageNet. $T$ is ResNet-50, $S_0$ is a ResNet-18, $R_{1}$ is a res-student whose architecture is searched by NAS. $S_1$ means using both $S_0$ and $R_1$. `SA' means using our sample-adaptive strategy at inference time. Energy is the metric we mention in \cref{ssub:res-learning}.
         \textbf{Bold}: The best results out of students. \ul{Underline}: The second best  results out of students.
        }
    \label{tab:tiny_image_final}
    \resizebox{1\textwidth}{!}{
    \begin{tabular}{lccccccc}
        \toprule
    Architecture & Optimizer & SA & Acc.~(\%) & \#Params (M) & MFLOPs & MFLOPs proportion to $T$ (\%) & Energy\\
    \midrule
    \emph{Teacher}  & & & & &\\
    Res50 & SGD &  & 70.42 / 75.50 & 23.92 / 25.56 & 4117 / 4119 & 100 / 100  &  0.9963 / 0.7782\\
    \midrule
    \emph{Students}  & & & & &\\
    $S_0$ & KD &  & 67.90 / 70.02 & 11.28 / 11.69 & 1821 / 1821 & \textbf{44.23} / \textbf{44.22}  &  0.9805 / 0.6575 \\
    
    \midrule
    \emph{ResKD students}  & & & & &\\
    $S_{1}$ & KD &  & \ul{68.98} / \textbf{73.73} & 16.38 / 19.71 & 2453 / 2777 & 59.58 / 67.42 & 0.9919 / 0.8040\\
    $S_{1}$ & KD & $\checkmark$ &  \textbf{68.99} / \ul{73.70} & 16.38 / 19.71 &  2206 / 2342 & \ul{53.59} / \ul{56.86} & -\\


    \bottomrule
    \end{tabular}
    }
    \end{center}
    \end{table*}

%% file: tables/L2_KL.tex
\begin{table}[tb]
\begin{center}
\caption{
    The effect of different KD loss functions $\mathcal{L}_{\mathrm{T-S}}$ on CIFAR-10. Res$X$-$Y$-$Z$ means we use ResNet-$X$ as $S_{0}$, ResNet-$Y$ as $R_{1}$, and ResNet-$Z$ as $R_{2}$. As a result, our framework is generic to $\mathcal{L}_{\mathrm{T-S}}$. \textbf{Bold}: The best results out of students. \ul{Underline}: The second best  results out of students.
}
\label{tab:L2_KL}
\resizebox{0.5\textwidth}{!}{
\begin{tabular}{lccccc}
    \toprule
Architecture & Optimizer & $\mathcal{L}_{KD}$ & Acc.~(\%) & \#Params (M) & MFLOPs\\
\midrule
\emph{Teacher}  & & & & &\\
Res110 & SGD & - & 94.19 & 1.728 & 255\\
\midrule
\emph{Students}  & & & & &\\
Res20 & SGD & - & 91.80 & 0.270     & \textbf{41}\\
Res20 & KD & $L_2$ & 93.06 & 0.270  & \textbf{41}\\
Res20 & KD & KL & 92.76 & 0.270     & \textbf{41}\\
Res20 & DML & $L_2$ & 92.53 & 0.270 & \textbf{41}\\
Res20 & RKD & $L_2$ & 92.45 & 0.270 & \textbf{41}\\
\midrule
\emph{ResKD students}  & & & & &\\
Res20-14 & KD & $L_2$ & 93.27 & 0.442       &  \ul{68}\\
Res20-14 & KD & KL & 93.50 & 0.442          &  \ul{68}\\
Res20-20 & KD & $L_2$ & 93.35 & 0.539 &  82\\
Res20-20 & KD & KL & 93.57 & 0.539 &  82\\
Res20-20-20 & KD & $L_2$ &  \textbf{93.95} & 0.809 &  123\\
Res20-20-20 & KD & KL &  \ul{93.93} & 0.809 &  123\\
Res20-14 & DML & $L_2$ & 92.88 & 0.442      &  \ul{68}\\
Res20-14 & RKD & $L_2$ & 93.17 & 0.442      &  \ul{68}\\
\bottomrule
\end{tabular}
}
\end{center}
\end{table} 

%% file: tables/backbone.tex
\begin{table}[tb]
\begin{center}
\caption{
    Performance on CIFAR-10 / CIFAR-100. $R_{1}$ is a res-student whose architecture is searched by NAS. The results of $S_1$ are combined with sample-adaptive inference. $S_1$ (VGG) / $S_1$ (Xception) / $S_1$ (Mobile) / $S_1$ (Shuffle) means $S_0$ is VGG11BN / Xception / MobileNetv2 / ShuffleNetv2 and $R_1$ is a NAS-based network. \textbf{Bold}: The best results out of students. \ul{Underline}: The second best  results out of students.
    }
\label{tab:backbone}

\begin{tabular}{lccc}
    \toprule
Architecture & Teacher & Acc.~(\%) & MFLOPs \\

\midrule
Inceptionv3 & -  & 95.28 / 79.40  & 3399 / 3400 \\
Xception & -  & 94.58 / 78.13 &1134 / 1134\\
Xception & Inceptionv3  & \ul{94.99} / \ul{78.58} & 1134 / 1134  \\
$S_1$ (Xception) & Inceptionv3  & \textbf{96.15} / \textbf{83.13}& 1163 / 1394  \\
\midrule
VGG19BN & -  & 93.70 / 71.50  &418 / 419\\
VGG11BN & -  & 91.99 / 68.96 & 172 / 173\\
VGG11BN & VGG19BN  & 92.17 / 71.46  &172 / 173  \\
$S_1$ (VGG) & VGG19BN  & 94.01  / 76.58  &206 / 203 \\
\midrule
ResNet50& - & 95.33 / 78.21  & 1305 / 1306  \\
MobileNetv2& - & 92.75 / 69.30 & 67 / 68 \\
MobileNetv2& ResNet50 &93.21 / 75.31& 67 / 68 \\
$S_1$ (Mobile)& ResNet50 & 93.83 / 78.30  & 70 / 101\\
\midrule
ShuffleNetv2& - &93.40 / 71.52 & 45 / 46  \\
ShuffleNetv2& ResNet50 &93.98 / 75.92 & \textbf{45} / \textbf{46}\\
$S_1$ (Shuffle)& ResNet50 &  94.56 / 78.57 & \ul{49} / \ul{65} \\
\bottomrule

\end{tabular}

\end{center}
\end{table}

%% file: tables/KD_method20.tex
\begin{table}[h]
    \begin{center}
    \caption{ The results of different distillation methods on CIFAR-100. $T$ is ResNet-110 with an accuracy of 74.31\% and a cost of 255.28 MFLOPs. $S_0$ is ResNet-20. $S_1$ is a NAS-based network $R_1$ combined with $S_0$.
    We report the accuracies as ``69.65 / 68.99 \cite{tian2019contrastive}'', which means that the re-implemented accuracy is 69.65, and the reported accuracy in \cite{tian2019contrastive} is 68.99.
    For a fair comparison, we control the cost of all $S_1$ at a similar level.
    Our ResKD can work well with different distillation methods.
    }
    \label{table.KD_method20main}
    %
    \begin{tabular}{llll}
        \toprule

    KD Method & Model &Accuracy (\%)&MFLOPs \\
    \toprule
    \multicolumn{4}{l}{\emph{Feature Distillation}} \\
    
    \multirow{2}{*}{FitNet \cite{romero2014fitnets}} & $S_0$ &69.65 / 68.99 \cite{tian2019contrastive}&41.22\\
                                 & $S_1$ &75.12 (\textcolor{red}{$+$5.47 / $+$6.13})&64.49 (\textcolor{red}{$+$23.27})\\
    \midrule
    \multirow{2}{*}{AT  \cite{zagoruyko2016paying}}     & $S_0$ & 70.50 / 70.22 \cite{tian2019contrastive} &41.22\\
                                 & $S_1$ &75.65 (\textcolor{red}{$+$5.15 / $+$5.43})&63.74 (\textcolor{red}{$+$22.52})\\
    \midrule
    \multirow{2}{*}{SP \cite{tung2019similarity}}     & $S_0$ & 70.72 / 70.04 \cite{tian2019contrastive} &41.22\\
                                 & $S_1$ &75.44 (\textcolor{red}{$+$4.72 / $+$5.40})&63.63 (\textcolor{red}{$+$22.41})\\
    \midrule
    \multirow{2}{*}{CC \cite{peng2019correlation}}     & $S_0$ & 69.44 / 69.48 \cite{tian2019contrastive} &41.22\\
                                 & $S_1$ &75.34  (\textcolor{red}{$+$5.90 / $+$5.86})&65.72 (\textcolor{red}{$+$24.50})\\
    \midrule
    \multirow{2}{*}{VID \cite{ahn2019variational}}    & $S_0$ & 70.22 / 70.16 \cite{tian2019contrastive} &41.22\\
                                 & $S_1$ & 76.10 (\textcolor{red}{$+$5.88 / $+$5.94})  &63.46 (\textcolor{red}{$+$22.24}) \\
    \midrule
    \multirow{2}{*}{RKD \cite{park2019relational}}    & $S_0$ & 69.40 / 69.25 \cite{tian2019contrastive} &41.22\\
                                 & $S_1$ &75.52 (\textcolor{red}{$+$6.12 / $+$6.27})&64.97 (\textcolor{red}{$+$23.75})\\
    \midrule
    \multirow{2}{*}{PKT \cite{passalis2018learning}}    & $S_0$ & 70.38 / 70.25 \cite{tian2019contrastive} &41.22\\
                                 & $S_1$ &75.82 (\textcolor{red}{$+$5.44 / $+$5.57})&64.02 (\textcolor{red}{$+$22.80})\\
    \midrule
    \multirow{2}{*}{AB \cite{heo2019knowledge}}     & $S_0$ & 69.94 / 69.53 \cite{tian2019contrastive} &41.22\\
                                 & $S_1$ &77.11 (\textcolor{red}{$+$7.17 / $+$7.58})& 64.75 (\textcolor{red}{$+$23.53})\\
    \midrule
    \multirow{2}{*}{FT \cite{kim2018paraphrasing}}     & $S_0$ &  70.06 / 70.22 \cite{tian2019contrastive} &41.22\\
                                 & $S_1$ &75.78 (\textcolor{red}{$+$5.72 / $+$5.56})& 64.45 (\textcolor{red}{$+$23.23})\\
    \midrule
    \multirow{2}{*}{NST \cite{huang2017like}}    & $S_0$ &  70.24 / 69.53 \cite{tian2019contrastive} &41.22\\
                                 & $S_1$ &75.93 (\textcolor{red}{$+$5.69 / $+$6.40})& 64.46 (\textcolor{red}{$+$23.24})\\
    \midrule
    \multirow{2}{*}{CRD \cite{tian2019contrastive}}    & $S_0$ &  71.33 / 71.46 \cite{tian2019contrastive} &41.22\\
                                 & $S_1$ &76.07  (\textcolor{red}{$+$4.74 / $+$4.61})& 62.80 (\textcolor{red}{$+$21.58})\\
    \toprule
    \multicolumn{4}{l}{\emph{Both Prediction and Feature Distillation}}\\
    \multirow{2}{*}{CRD+KD \cite{tian2019contrastive}} & $S_0$ &71.61 /  71.56 \cite{tian2019contrastive} &41.22\\
                                 & $S_1$ & 75.75  (\textcolor{red}{$+$4.14 / $+$4.19})& 57.37 (\textcolor{red}{$+$16.15})\\
    \bottomrule%
    
    \end{tabular}
    
    \end{center}
    \end{table}

%% file: sections/analysis.tex
\section{Analysis: Why Residual-Guided Learning Works}
In this section, we try to shed some light on why and how our residual-guided learning helps the training process.

\subsection{Theoretical Analysis}

\input{tables/gi}
Our residual-guided learning is based on the gap between a teacher $T$ and a student $S$. We should use a suitable metric to measure the gap between $T$ and $S$. Many metrics are already proposed from simple $L_1$, $L_2$ to complex metrics in \mbox{\cite{hu2015deep, zhang2019adasample, ding2016robust, 7574389}}.
Inspired by \cite{zhang2019adasample}, we measure the informativeness of training examples by analyzing their resulting gradients since the training data contribute to optimization via gradients. 
The gap of informativeness (gap\_info, GI)  between $S$ and $T$ for a training example $\mathbf{x}^{(j)}$ at an iteration $t$   is defined as:

\begin{equation}
\begin{split}
\label{eq:info}
\mathrm{GI}(\mathbf{x}^{(j)},S, T, t) &=\|\nabla_{\theta_{t}}\mathcal{L}(S(\mathbf{x}^{(j)}),T(\mathbf{x}^{(j)}))\|_2,
\end{split}
\end{equation}
where $\nabla_{\theta_t}$ denotes the gradients of $S$'s parameters $\theta$ at the iteration $t$.
However, computing this $L_2$-norm directly is expensive.
Instead, we could estimate the upper bound $\widehat{\mathrm{GI}}$ for $\mathrm{GI}$. 

Following the work of \cite{katharopoulos2018not} and without loss of generality,  we use a multi-layer perceptron (MLP) as the model in our analysis.
Let $\theta^{(l)}\in\mathcal{R}^{M_{l}\times M_{l-1}}$ be the weight matrix for layer $l$ and $\sigma^{(l)}(\cdot)$ be a Lipschitz continuous activation function, and then we have:

\begin{equation}
\begin{aligned}
\label{eq:defmlp}
&\mathbf{a}^{(0)} = \mathbf{x}^{(j)},\\
&h^{(l)} = \theta^{(l)}\mathbf{a}^{(l-1)},\\
&\mathbf{a}^{(l)} = 	\sigma^{(l)}(h^{(l)}),\\
&f(\mathbf{x}^{(j)},\theta) = \mathbf{a}^{(L)}, \\
\end{aligned}
\end{equation}
where $\mathbf{a}^{(l)}$ denotes the feature maps after layer $l$ and $\mathbf{x}^{(j)}$ is a certain sample. We define:
\begin{equation}
\begin{split}
\label{eq:dig}
	\Sigma^{'}_{l}(h^{(l)}) = \mathrm{diag}(\sigma^{'(l)}(h^{(l)}_{1}),\sigma^{'(l)}(h^{(l)}_{2}) \cdots,\sigma^{'(l)}(h^{(l)}_{M_{l}})). 
\end{split}
\end{equation}

\begin{equation}
\begin{split}
\label{eq:pi}
\Pi^{(l)} = (\prod_{i=l}^{L-1}\Sigma^{'}_{i}(h^{(i)})\theta^{T}_{i+1})\Sigma^{'}_{L}(h^{(l)}). 
\end{split}
\end{equation}
$\mathcal{L}$ is the loss function $\mathcal{L}(S(\mathbf{x}^{(i)}),T(\mathbf{x}^{(i)}))$ in \cref{eq:info}. 
The $\mathrm{GI}^{(l)}$ is the informativeness of the parameters in layer $l$ and it can be expressed as:
\begin{equation}
\begin{aligned}
\label{eq:first}
	\mathrm{GI}^{(l)} &= \|(\Pi^{(l)} \nabla_{\mathbf{a}^{(L)}} \mathcal{L})(\mathbf{a}^{(l-1)})^{T}\|_2\\
&\le \|\Pi^{(l)}\|_2 \|(\mathbf{a}^{(l-1)})^{T}\|_2\|\nabla_{\mathbf{a}^{(L)}} \mathcal{L}\|_2. 
\end{aligned}
\end{equation}
Various weight initialization \cite{glorot2010understanding} and activation normalization techniques \cite{ioffe2015batch}, \cite{ba2016layer} uniformize the activations across samples. 
As a result, the variation of the gradient norm is mostly captured by the gradient of the loss function with respect to the pre-activation outputs of the last layer of our neural network. 
Consequently, we can derive the following upper bound to the gradient norm of all the parameters. 
Suppose that $C_{max}$ is a constant:
\begin{equation}
\begin{split}
\label{eq:C}
	\mathrm{GI} \le C_{max}\|\nabla_{\mathbf{a}^{(L)}} \mathcal{L}\|_2. 
\end{split}
\end{equation}
Based on \cref{eq:C}, we set:
\begin{equation}
\begin{split}
\label{eq:C_hat}
	\widehat{\mathrm{GI}} = C_{max}\|\nabla_{\mathbf{a}^{(L)}} \mathcal{L}\|_2. 
\end{split}
\end{equation}

For our method, we use $L_2$ distance loss function to measure the difference between current student network and the guidance of current teacher:
\begin{equation}
\begin{aligned}
\label{eq:LKD}
&\mathcal{L} = \|S - T\|_2^2. 
\end{aligned}
\end{equation}
We calculate the first derivative of our loss function:
\begin{equation}
\begin{split}
\label{eq:gKD}
\nabla_{\mathbf{a}^{(L)}} \mathcal{L} =  2 \cdot \|S-T\|_2. 
\end{split}
\end{equation}

According to \cref{eq:C_hat} and \cref{eq:gKD}, we set $C_1 = 2C_{max}$, and we define $S_0^* / R_i^*$ is the result that the $S_0 / R_i$ has been optimized well:
\begin{equation}
\begin{aligned}
\label{eq:upper2}
	&\widehat{\mathrm{GI}}_{S_{0}} = C_{1}\|S_{0}-T\|_2, \\
	&\widehat{\mathrm{GI}}_{S_{1}}= C_{1}\|R_{1}-(T-S_{0}^*)\|_2, \\
	&\widehat{\mathrm{GI}}_{S_{2}} = C_{1}\|R_{2}-(T-S_{0}^*-R_{1}^*)\|_2, \\
&\cdots. 
\end{aligned}
\end{equation}
With proper optimization for $R_{1}$, we have the best $R_1^*$. 
The best $R_1^*$ is better than other value of $R_1$ include that $R_1$ always equals $0$, so we have the upper bound of $ \widehat{\mathrm{GI}}_{S_{1}}$ in residual-guided knowledge distillation. 
\begin{equation}
\begin{aligned}
\label{eq:upper3}
	\widehat{\mathrm{GI}}_{S_{1}} &\le C_{1}\|0-(T-S_{0}^*)\|_2\\
	&\le \widehat{\mathrm{GI}}_{S_{0}}. 
\end{aligned}
\end{equation}
We rewrite Eq. \eqref{eq:upper3} and get similar conclusion of other equations, and set $\Delta^{(\mathrm{GI})}_{i} \ge 0$:
\begin{equation}
\begin{aligned}
\label{eq:upper4}
	&\widehat{\mathrm{GI}}_{S_{0}} = \widehat{\mathrm{GI}}_{S_{1}} + \Delta^{(\mathrm{GI})}_{0}, \\
	&\widehat{\mathrm{GI}}_{S_{1}} = \widehat{\mathrm{GI}}_{S_{2}} + \Delta^{(\mathrm{GI})}_{1}, \\
&\cdots, \\
	&\widehat{\mathrm{GI}}_{S_{n-1}} = \widehat{\mathrm{GI}}_{S_{n}} + \Delta^{(\mathrm{GI})}_{n-1}. \\
\end{aligned}
\end{equation}
We add all the equations in \cref{eq:upper4}:
\begin{equation}
\begin{aligned}
\label{eq:upper5}
	&\widehat{\mathrm{GI}}_{S_{0}} = \widehat{\mathrm{GI}}_{S_{n}} + \sum_{i=0}^{n-1}\Delta^{(\mathrm{GI})}_{i}. \\
\end{aligned}
\end{equation}

$R_i^*$ is the best result of optimizing $R_i$ to approach $T - S_0 - \sum_{j=1}^{i-1} R_j$, so we suppose that when we have optimized $R_i^*$, $\widehat{\mathrm{GI}}_{s_i}$ has changed to $k_i(T - S_0 - \sum_{j=1}^{i-1}R_j)$ and $k_i \in (0,1)$. 
In this constraint, $\widehat{\mathrm{GI}}_{s_i}$ can be rewritten as:
\begin{equation}
\begin{aligned}
\label{eq:k1}
	&\Delta^{(\mathrm{GI})}_{i} = k_i\cdot\widehat{\mathrm{GI}}_{S_{i-1}}, k_i \in (0,1). \\
\end{aligned}
\end{equation}
When the $R_i$ becomes stronger, the $k_i$ becomes larger and the network $R_i$ bridge the gap better. 
Also the $\widehat{\mathrm{GI}}_{S_{i}}$ can be rewritten as:
\begin{equation}
\begin{aligned}
\label{eq:k2}
	\widehat{\mathrm{GI}}_{S_{i}} = &(\prod_{j=1}^{i}(1-k_j))(T-S_0^*),\\
 &k_j \in (0,1). \\
\end{aligned}
\end{equation} 
We can learn that the final performance of $S_i$ is depended on the expression ability of each network in $S_0, R_1, \cdots, R_n$.
In \cref{ssec:Residualguided}, we will show some results that how the choice of a certain res-student network and the number of res-student networks affect the final performance.

\begin{figure}[tb]
	\begin{subfigure}{0.33\linewidth}
		\centering
		\includegraphics[width=0.95\linewidth]{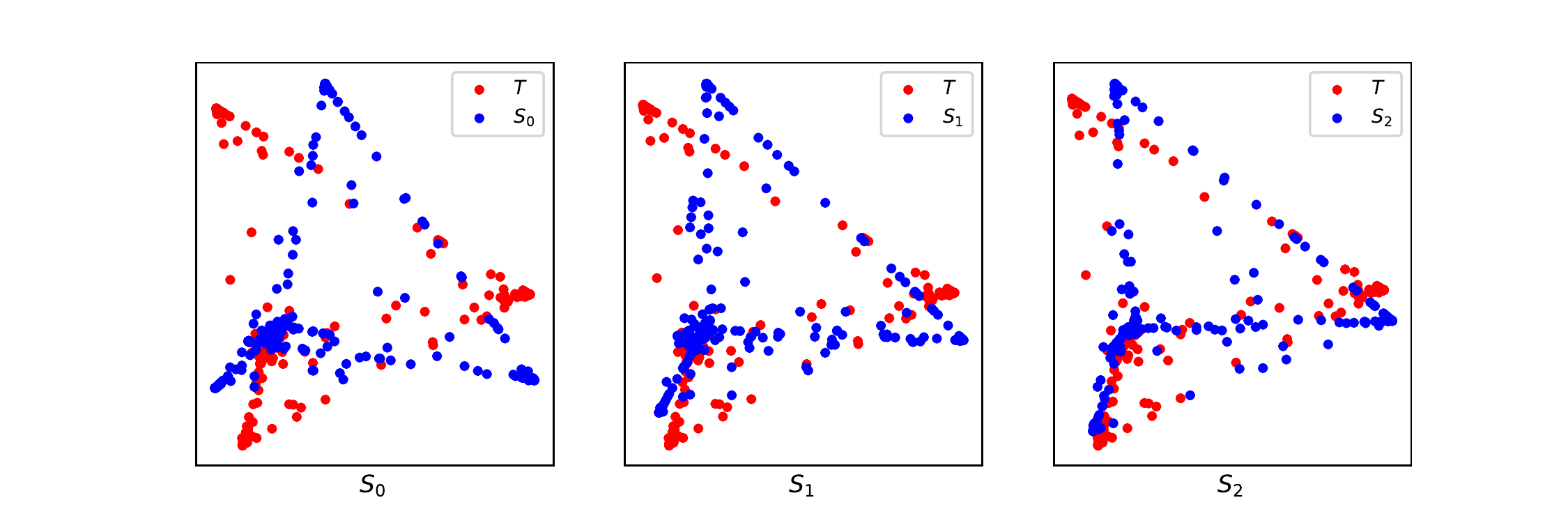}\\
		\caption{$S_0$} \label{anal_s0}
	\end{subfigure}%
	\begin{subfigure}{0.33\linewidth}
		\centering
		\includegraphics[width=0.95\linewidth]{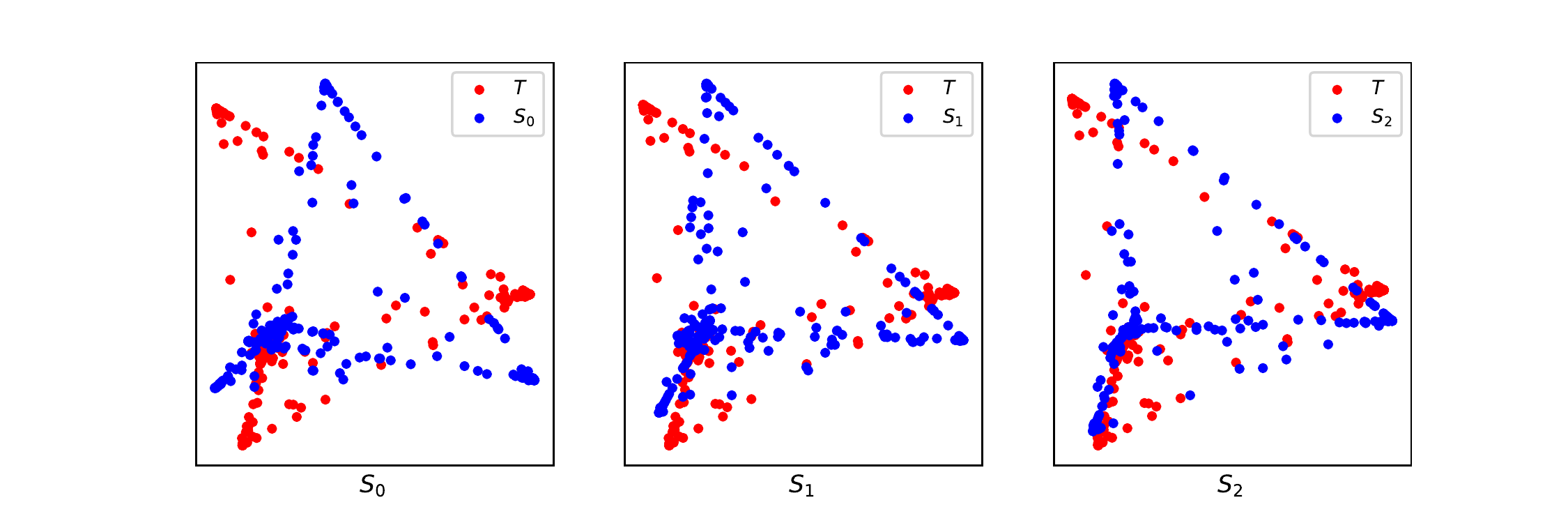}\\
		\caption{$S_1$} \label{anal_s1}
	\end{subfigure}%
	\begin{subfigure}{0.33\linewidth}
		\centering
		\includegraphics[width=0.95\linewidth]{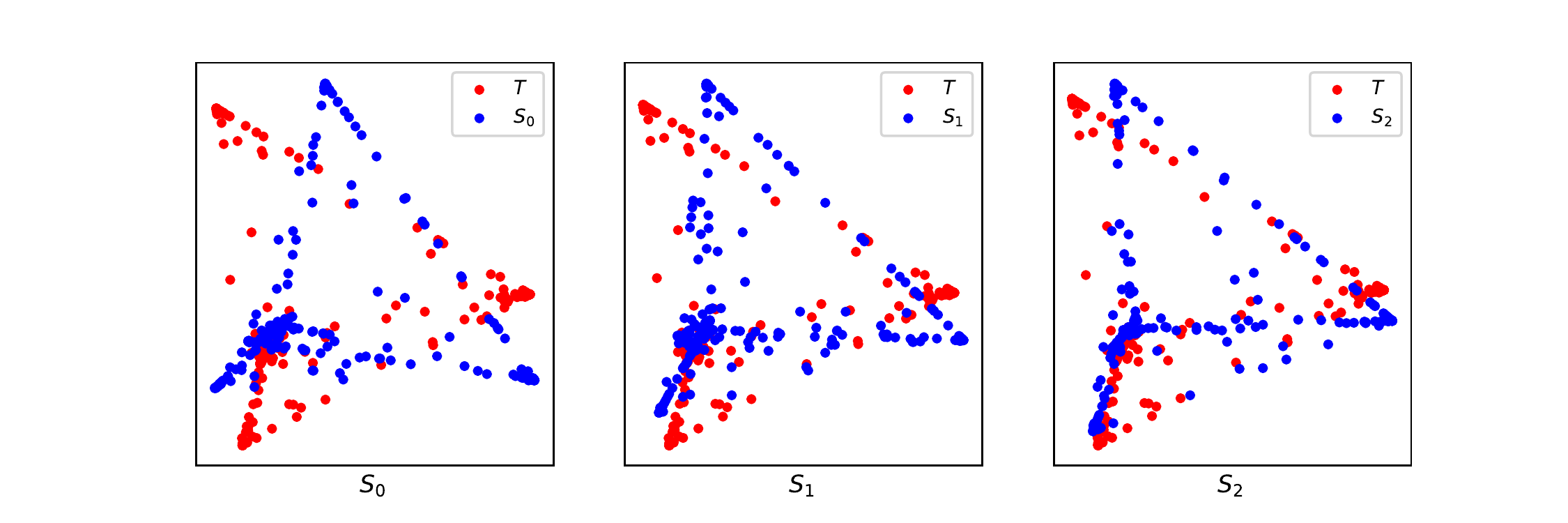}\\
		\caption{$S_2$} \label{anal_s0}
	\end{subfigure}%
	\centering
	\caption{The visualization of 2D PCA representation of ResKD student's logits using different number of res-students on CIFAR-10. The red dots in (a), (b) and (c) are the distribution of $T$, ResNet-110, and they are the same one. (a) the distribution of $S_0$. (b) the distribution of $S_1$. (c) the distribution of $S_2$. }
	\label{fignum}
\end{figure}

\subsection{Empirical Analysis}

In this part, we show the $\mathrm{GI}$ of different students to validate our theoretical analysis on CIFAR-10 first. 
Next, we show the distribution of logits of different students directly to validate our residual-guided knowledge distillation.

\subsubsection{Gap\_info (GI) Observation} We empirically verify whether the expression ability of the student network and res-student networks will influence the $\mathrm{GI}$. 
The average $\mathrm{GI}$ and the average logits distribution samples that anyone of our residual-guided model predicts correctly and $S_0$ gives the wrong prediction are showed. 
We use ResNet-110 as the teacher network and ResNet-20 as the $S_0$ on CIFAR-10. 
When we focus on $R_2$, we set ResNet-20 as $R_1$ and observe how different res-student network $R_2$ affects the final performance.
The expression of Res$X$-$Y$-$Z$ means that we use ResNet-$X$ as $S_{0}$, use ResNet-$Y$ as $R_{1}$ and use ResNet-$Z$ as $R_{2}$.

\begin{figure}[tb]
		\begin{subfigure}{0.33\linewidth}
			\centering
			\includegraphics[width=0.95\linewidth]{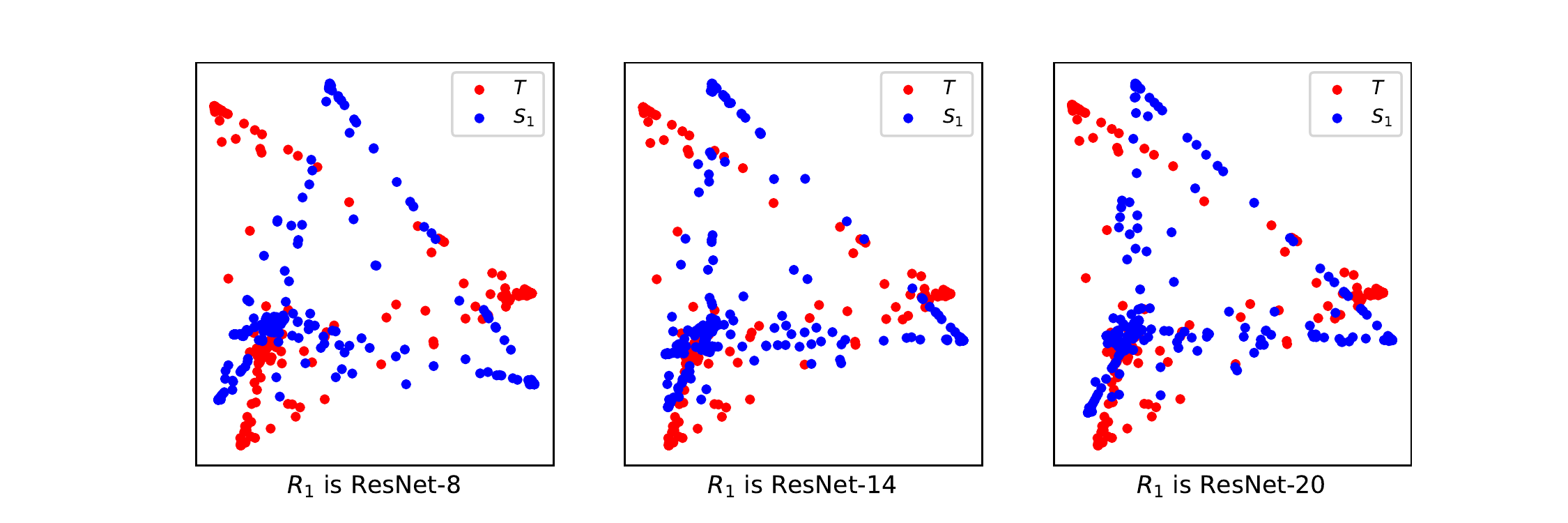}\\
			\caption{$R_1$ is ResNet-8} \label{anal_s0}
		\end{subfigure}%
		\begin{subfigure}{0.33\linewidth}
			\centering
			\includegraphics[width=0.95\linewidth]{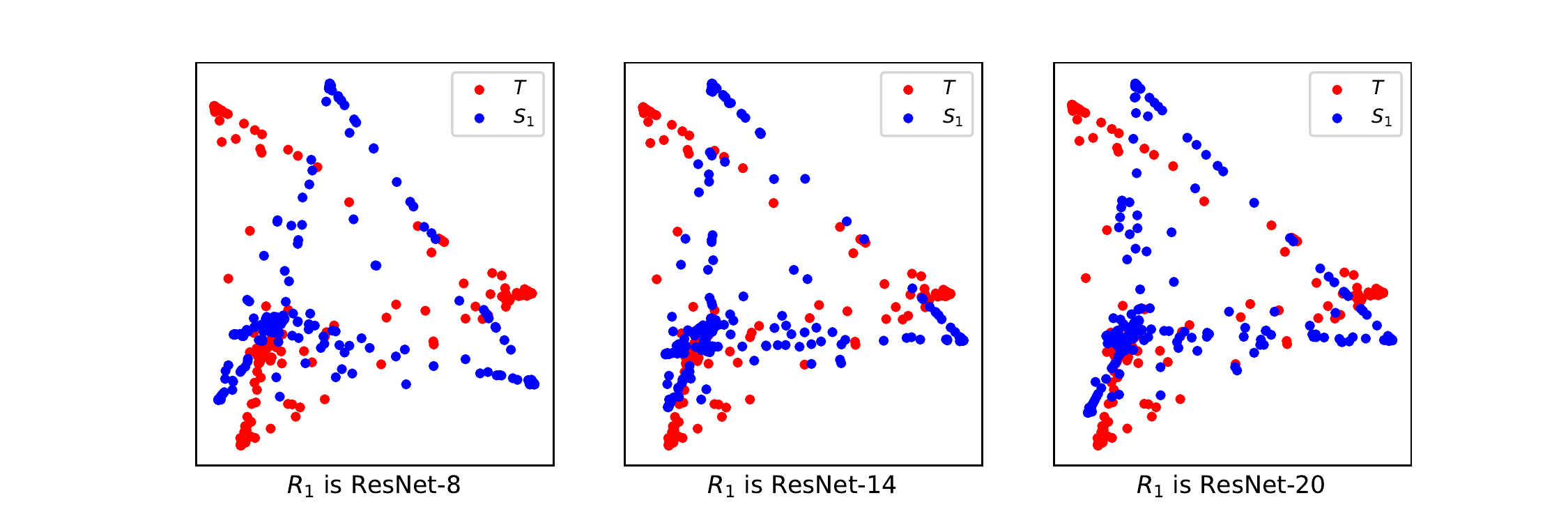}\\
			\caption{$R_1$ is ResNet-14} \label{anal_s1}
		\end{subfigure}%
		\begin{subfigure}{0.33\linewidth}
			\centering
			\includegraphics[width=0.95\linewidth]{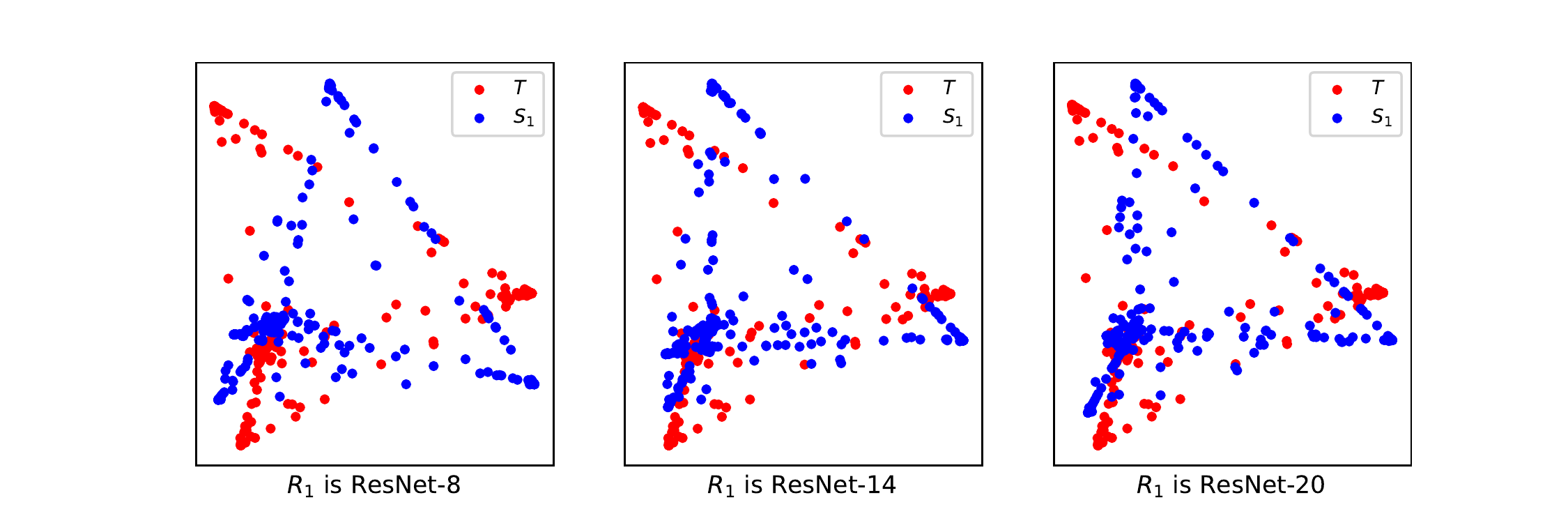}\\
			\caption{$R_1$ is ResNet-20} \label{anal_s0}
		\end{subfigure}%

		\begin{subfigure}{0.33\linewidth}
			\centering
			\includegraphics[width=0.95\linewidth]{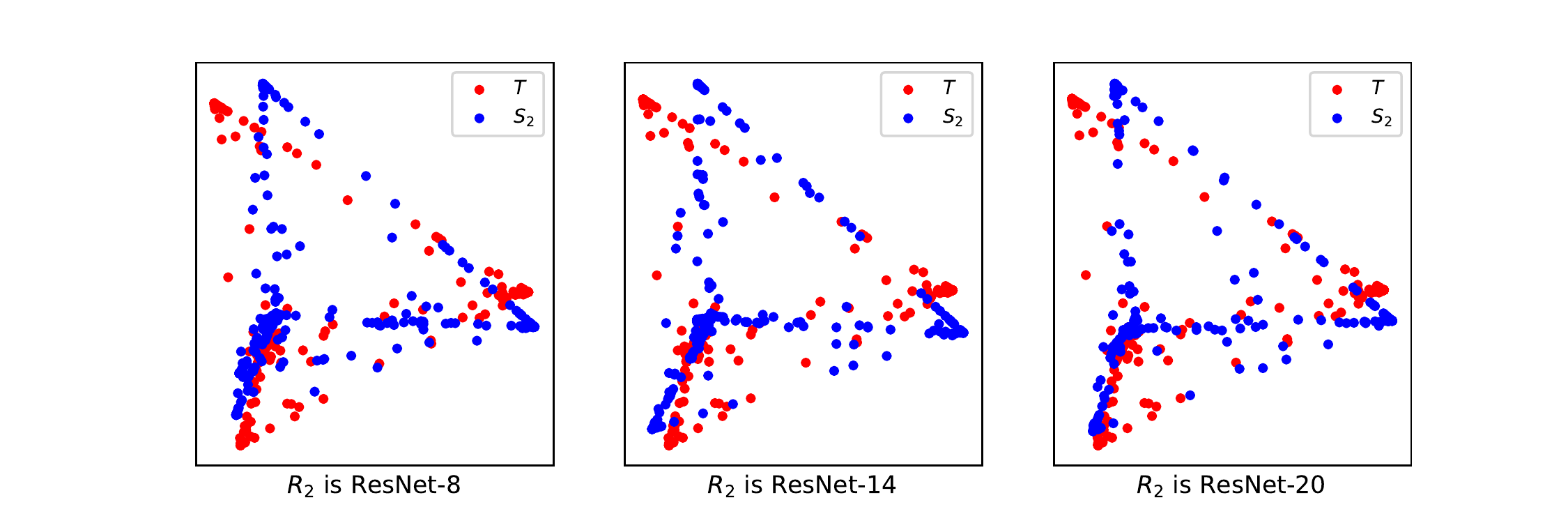}\\
			\caption{$R_2$ is ResNet-8} \label{anal_s0}
		\end{subfigure}%
		\begin{subfigure}{0.33\linewidth}
			\centering
			\includegraphics[width=0.95\linewidth]{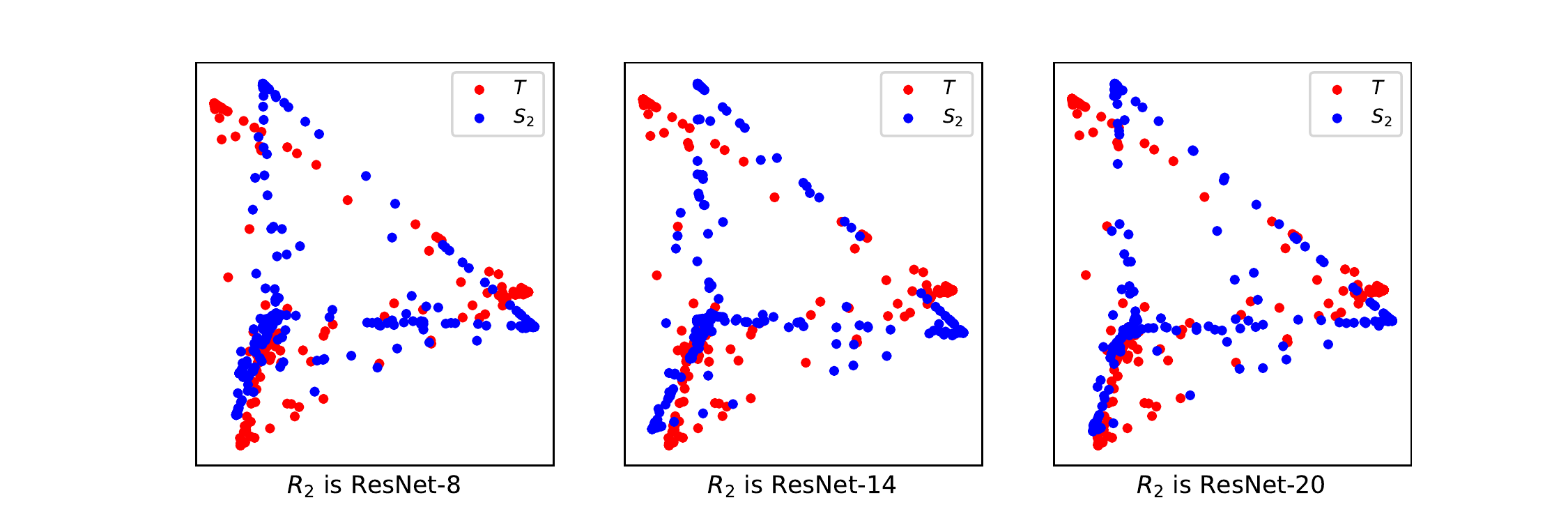}\\
			\caption{$R_2$ is ResNet-14} \label{anal_s1}
		\end{subfigure}%
		\begin{subfigure}{0.33\linewidth}
			\centering
			\includegraphics[width=0.95\linewidth]{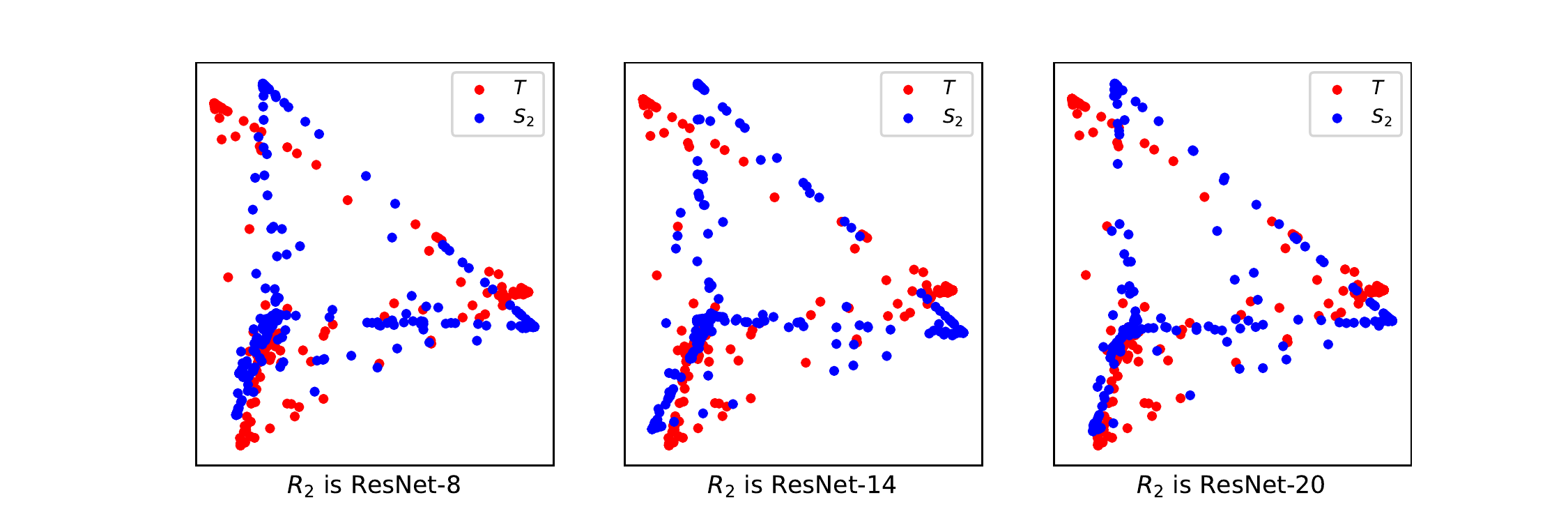}\\
			\caption{$R_2$ is ResNet-20} \label{anal_s0}
		\end{subfigure}%
		\centering
		\caption{The visualization of 2D PCA representation of $S_1$'s / $S_2$'s logits using different $R_1$ / $R_2$ on CIFAR-10. 
		The red dots are the distribution of $T$, ResNet-110, and they are the same one in (a), (b), (c), (d), (e), and (f). 
		The blue dots in (a), (b), and (c) are the distributions of different $S_1$ using different $R_1$. $S_0$ is ResNet-20. 
		The blue dots in (d), (e), and (f) are the distributions of $S_2$ using different $R_2$ and using ResNet-20 as both $S_0$ and $R_1$.
		}
		\label{figr12}
	\end{figure}

Firstly, we discuss the influence when the number of res-student networks used increases. 
The results are shown in \cref{tab:gi}. 
When we use the first res-student network $R_1$ to bridge the gap between $T$ and $S_0$, the output of $S_0 + R_1$ is more similar to the output of $T$ (the average second norm to T is less) and the $\mathrm{GI}$ is smaller than the one of $S_0$. 
Similarly, when we use $R_2$, the output of $S_0+R_1+R_2$ is more likely to $T$ and the $\mathrm{GI}$ continues decreasing. 
When we use ResNet-20 for $S_0$, $R_1$ and $R_2$, the $\mathrm{GI}$ is about the half of $S_0$.

Next, we discuss the difference among different latest res-student networks. In \cref{tab:gi}, the different number of res-student networks can cause different results. 
When we use ResNet-8 / 14 / 20 to for $R_1$, the ``$\mathrm{GI}/C_1$ to $T$'' and ``$L_2$ to $T$'' decrease, and when we use ResNet-20, the value is the smallest. The phenomenon is similar when we use ResNet-8 / 14 / 20 to for $R_2$ and fix $S_0$ and $R_1$. We can learn that the stronger $R_1 / R_2$ is, the more similar $S_1 / S_2$ is with the $T$.

\subsubsection{Logits Visualization}
We show the 2D PCA representation of different $S_i$'s logits on CIFAR-10. 
In each figure, the red points are the same one and indicate the distribution of $T$'s logits.

In \cref{fignum}, the blue points indicate the distributions of $S_0$'s, $S_1$'s and $S_2$'s logits.
We use ResNet-20 for $S_0$, $R_1$ and $R_2$ in this figure. 
We can learn that with the number of res-students increasing, the distribution of $S_i$'s logits is more and more similar to the distribution of $T$'s. 

In \mbox{\cref{figr12}}, the blue points indicate the distribution of $S_1$ / $S_2$. 
We use ResNet-20 / 14 / 8 for $R_1$ / $R_2$, $S_0$ is ResNet-20, and $R_1$ is ResNet-20 when we focus $R_2$. 
We can learn that when res-students $R_1$ / $R_2$ is stronger, the distribution of $S_1$ / $S_2$ is more similar with the distribution of $T$.

%% file: tables/gi.tex
%

\begin{table}[tb]
\begin{center}
	\caption{$\mathrm{GI}$ of different student networks to the teacher network on CIFAR-10. $T$ is ResNet-110, and $S_0$ is ResNet-20. Res$X$-$Y$-$Z$ means that we use ResNet-$X$ as $S_0$, ResNet-$Y$ as $R_1$, and ResNet-$Z$ as $R_2$.
\textbf{Bold}: The best results out of students.
 	}
\label{tab:gi}
\begin{tabular}{lccc}
\toprule
	Architecture & Acc.~(\%) & $L_2$ to T & $\mathrm{GI}/C_1$ to T \\
\midrule
\emph{Teacher}  & & & \\
Res110 & 94.19 & - & - \\
\midrule
\emph{Student }  & & & \\
Res20 & 93.06 & 0.90 & 1.73\\
\midrule
\emph{ResKD networks}  & & & \\
Res20-8     & 93.03 & 0.82 & 1.64\\
Res20-14    & 93.27 & 0.74 & 1.53\\
Res20-20    & 93.35 & 0.61 & 1.37 \\
Res20-20-8  & 93.55 & 0.49 & 1.15\\
Res20-20-14 & 93.84 & 0.43 & 1.03\\
Res20-20-20 &  \textbf{93.95} &  \textbf{0.37} &  \textbf{0.89}\\
\bottomrule
\end{tabular}
\end{center}
\end{table}

%% file: sections/conclusion.tex
\section{Conclusion}
We have studied an under-explored yet important field in knowledge distillation of neural networks. 
We have shown that using res-students to bridge the gap between student and teacher is a key to improve the quality of knowledge distillation. We propose our residual-guided learning and sample-adaptive inference to realize this idea. 
We also validate the effectiveness of our approach in various datasets and studied its properties both empirically and theoretically. 